\documentclass[lettersize,journal]{IEEEtran}
\usepackage{amsmath,amsfonts}
\usepackage{algorithmic}
\usepackage{algorithm}
\usepackage{array}
\usepackage{wrapfig}
\usepackage[caption=false,font=normalsize,labelfont=sf,textfont=sf]{subfig}
\usepackage{textcomp}
\usepackage{stfloats}
\usepackage{url}
\usepackage{verbatim}
\usepackage{graphicx}
\usepackage{booktabs}
\usepackage{cite}
\usepackage{multirow}
\usepackage{color, soul}
\usepackage{balance}
\usepackage{xcolor}

\begin{document}
\bstctlcite{BSTcontrol}
\title{PEARL: A Lightweight Prompt-based Feature Interpreter Framework for Real-Time, Anonymous, and Heterogeneous Collaborative Perception}

\author{Armin Maleki, \textit{Graduate Student Member, IEEE}  and Hayder Radha, \textit{Fellow, IEEE}\\
Department of Electrical and Computer Engineering\\ 
Michigan State University\\
{\tt\small malekiar@msu.edu},
{\tt\small radha@msu.edu}
}

\maketitle

\begin{abstract}
Heterogeneity across Collaborative Perception (CP) agents is a major challenge for emerging CP frameworks due to domain gaps from differing sensors, architectures, and training data. Prior works mitigate this challenge by aligning features in a unified space via model retraining or per-agent-type interpreters. These strategies (a) require access to neighbor configurations, (b) do not fully address realistic real-time CP deployment, and (c) generalize poorly to unseen agents joining at run time. To overcome these challenges, we present \textbf{PEARL}, a Prompt-Embedding framework for Anonymous and Real-time Lightweight heterogeneous CP. PEARL supports multiple CP interpreters, one of which can be optimally selected for a new-joining agent in real-time by aligning intermediate BEV features using two lightweight, multi-scale interpreters trained in parallel: a sparse-detection (LWSD) interpreter that aligns salient regions for cooperative detection, and a dense, domain-invariant (LWDDI) interpreter that produces agent-invariant features for fast interpreter selection in real-time. To keep computation and storage low, both interpreters use low-rank visual prompts, significantly reducing model complexity. Extensive experiments on simulated (OPV2V, V2XSet) and real (DAIR-V2X) datasets show that PEARL generalizes consistently across simulated and real-world cooperative driving scenarios, and its real-time model selection strategy yields an 8.2\% Average Precision (AP) gain over a random-selection baseline while running in 1.67 ms on average. 
Although primarily designed for real-time CP, PEARL also outperforms state-of-the-art heterogeneous CP frameworks under traditional offline training by 5.6\% in AP, on average, while reducing communication cost by up to 34.7×. Equally important, PEARL does not require the sharing of agents' configurations or models' settings, and hence it protects information that agents may deem as proprietary or private. All of these benefits and results establish PEARL as a scalable and practical framework for heterogeneous collaborative perception.
\textnormal {Code will be available at~\url{https://github.com/arminmaleki007/PEARL}.}

\end{abstract}


\section{Introduction}
\label{sec:intro}

Advances in collaborative perception (CP) have the potential to accelerate the deployment of autonomous vehicles and substantially improve safety~\cite{huang2023v2x,yazgan2024survey}. By aggregating observations across multiple agents, CP provides each vehicle with a broader, less-occluded field of view and improves detection and decision time over single-vehicle perception. However, practical deployment faces bandwidth limitations, communication noise, packet loss, and less-explored heterogeneity across agents. While prior works address pose error~\cite{song2024spatial,ni2024self,lei2024robust}, bandwidth~\cite{hu2024communication,hu2022where2comm,yang2023what2comm}, and latency~\cite{lei2024robust,wang2024intercoop,zheng2024cooperfuse}, they typically assume identical cooperating agents. This assumption is unrealistic: manufacturers use diverse sensor suites and models, and even when architectures are matched, different datasets or training yield different weights, leading to mismatched intermediate features. Explicitly modeling and mitigating \textbf{heterogeneity} is therefore critical for the safe, scalable, and reliable deployment of collaborative perception.

Prior heterogeneous CP efforts follow two strategies for bridging inter-agent feature domain gaps toward a unified feature space: (i) model retraining and (ii) per-agent-type interpreters. In (i), frameworks standardize on a homogeneous space (often the ego’s) and retrain the collaborative stack or neighbor submodules~\cite{lu2024extensible,kong2025cobevmoe,shao2025negocollab,li2024di,xiang2023hm,xu2022v2x} to ensure compatibility; this is typically slow for large stacks, raises privacy risks (sharing sensor/model details), and can degrade single-vehicle performance. In (ii), per-type interpreters map features into a shared space~\cite{xu2022bridging,xin2025pnpda+,xia2025one,lu2025privacy}; while avoiding exposure of sensor/model configurations, they increase model complexity and may incur larger performance degradation than retraining. 
Despite their merits, both directions struggle to generalize efficiently to unseen agent types while simultaneously maintaining accuracy, preserving privacy, and remaining computationally feasible for real-time deployment. Moreover, practical heterogeneous CP should generalize across different datasets and sensing settings, including simulated and, more importantly, real datasets.

When the ego agent maintains several offline-trained heterogeneous CP models/interpreters and must select one at inference for a newly joining agent, three challenges arise. (i) \emph{metadata dependence and privacy}\footnote{In this paper, the notion of \textit{privacy} is in the context of \textit{proprietary information or data} that an agent may not be willing to share, including types and resolutions of sensors used by the agent, particular object detection models, and related parameters/hyperparameters of these models.}—selection typically requires the joining agent’s type, sensor, and model configuration, forcing disclosure of sensitive metadata that the joining agent could consider proprietary or private. (ii) \emph{semantic mismatch within a nominal type}—agents may produce features with different semantics than those seen during training due to data, model, or training stochasticity, which degrades overall performance. (iii) \emph{previously unseen agent types}—types not covered by offline training may not efficiently align at join time. Therefore, viable CP must admit new agents while minimizing privacy exposure and accuracy loss in real-time; rejecting agents, requiring sensitive disclosures, or naïve selection, e.g., randomly picking a model, undermines reliability, generalization, and safety, where online detection testing across multiple samples requires ground truth and is too slow for real-time operation.

Realistic collaborative perception systems may operate in highly dynamic traffic environments where multiple neighboring agents join or leave simultaneously. In practice, a CP agent needs to accommodate and support a rather large number of potential agent types and models, reflecting the realistic heterogeneity and diversity of other CP agents. Moreover, CP frameworks are developed to improve perception over single-agent operation; however, if the selected interpreter is poorly matched or the heterogeneous features are not properly aligned, the collaborative model may perform worse than the agent’s own \textit{single-agent} model. Therefore, interpreter assignment \textit{in real-time} must remain accurate and efficient as the number of agent types and candidate interpreters increases. These challenges motivate a runtime selection mechanism that is both \textit{scalable} and \textit{reliable}: it should maintain low wall-clock latency as the interpreter pool grows while selecting interpreters that preserve the expected benefit of collaboration.

In summary, and to the best of our knowledge, truly heterogeneous and scalable CP has not been addressed under realistic and real-time constraints. To close this major gap in CP research, we propose \textbf{PEARL}—a \textbf{P}rompt-\textbf{E}mbedding framework for \textbf{A}nonymous and \textbf{R}eal-time \textbf{L}ightweight heterogeneous CP. \textbf{PEARL} consists of two stages and two parallel pipelines, trained to preserve a joining agent’s sensor/model anonymity while remaining computationally efficient. Both pipelines employ multi-scale, coarse-to-fine interpreters to align a new agent’s Bird-Eye-View (BEV) features to the ego space: in Stage-1, pipeline-1 uses LightWeight Sparse-Detection (\textbf{LWSD}) prompts with cross-attention and 3D spatial attention for downstream cooperative detection, while pipeline-2 uses LightWeight Dense, Domain-Invariant (\textbf{LWDDI}) prompts with an adversarial domain classifier to produce agent-type-invariant features compatible with the ego space, forming a heterogeneous collaboration base. In Stage-2, we fine-tune (a) $\mathrm{LWSD}$ from pipeline-1 and small detection utilities (foreground estimator and multi-scale extractor), and (b) $\mathrm{LWDDI}$ and its classifier from pipeline-2 to yield updated domain-invariant features. This design maps new agents’ features to the ego space without exposing metadata, preserves single-vehicle accuracy and privacy, avoids costly retraining/storage, and supports scalable deployment.

At runtime, after offline training on multiple agent types, the ego vehicle runs each pretrained $\mathrm{LWDDI}$ path to produce domain-invariant features from the joining agent’s intermediate features. It then assigns the agent to the highest-similarity candidate by computing a similarity measure with the ego’s own features, thereby constructing a domain-invariant space around the ego’s BEV for efficient real-time interpreter selection. Otherwise, agent-specific features from other pre-trained heterogeneous agents would not be available at deployment; producing them would require running their encoders, revealing sensor/model metadata, and incurring substantial computation. Because this process operates on domain-invariant intermediate features rather than running a full detector, such a selection approach runs much faster than detection in real-time and requires no disclosure of the joining agent’s sensor or model metadata. Moreover, embeddings from different model/weight variants often share semantics, enabling assignment of an available model to an unseen agent by leveraging these shared semantics. Taken together, these choices advance privacy-preserving, real-time heterogeneous CP.

To efficiently construct and train our interpreters, i.e., $\mathrm{LWSD}$ and $\mathrm{LWDDI}$, we use visual prompts, learnable tensors injected into intermediate features. $\mathrm{LWSD}$ uses \emph{sparse} prompts to emphasize salient regions, improving alignment of detection-critical features and reducing the detection domain gap; $\mathrm{LWDDI}$ uses \emph{dense} prompts that capture holistic domain cues and yield agent-invariant representations under adversarial training, which minimizes the distance between a new agent’s intermediate features and the ego feature space. Since prompts are 3D tensors, naïve prompting across both pipelines is computationally expensive, so we adopt a \emph{low-rank} prompt factorization~\cite{carroll1970analysis} that reduces trainable parameters and computational complexity while retaining alignment capacity and downstream performance. Critically, low-rank prompts in the domain-invariant path reduce feature-scoring cost, enabling selection via a single aggregated cosine similarity for each candidate, and meeting real-time privacy conditions.

A preliminary version of this work appeared in CVPR 2026 Findings~\cite{maleki2026pearl}. This version substantially extends the conference paper by adding: (a) evaluations on V2XSet and real-world DAIR-V2X datasets, (b) cross-dataset interpreter selection, (c) scalable multi-interpreter deployment analysis, (d) new model-selection baselines, (e) CP communication bandwidth analysis,
and (f) extra ablation studies. Overall, the key contributions of this paper can be summarized as follows:
\begin{itemize}

\item \textbf{Prompt-Embedding framework for Anonymous and Real-time Lightweight CP.} We introduce a new framework for real-time, heterogeneous, and anonymous CP 
which exploits coarse-to-fine, multi-scale features to bridge feature-domain gaps among heterogeneous agents with two complementary pipelines.

\item \textbf{PEARL's lightweight interpreters and multi-scale prompts.} Lightweight visual-prompt interpreters in both pipelines are introduced: $\mathrm{LWSD}$ with \emph{sparse} prompts to align detection-critical features, and $\mathrm{LWDDI}$ with \emph{dense} prompts to learn agent-invariant representations via adversarial training. A \emph{low-rank} factorization of multi-scale prompts is used along with these interpreters to reduce computational complexity, enabling privacy-preserving (anonymous), real-time model assignment using a single aggregated similarity score on $\mathrm{LWDDI}$ features.

\item \textbf{Efficient adaptation to unseen heterogeneous agents.}
We develop a two-stage training strategy that efficiently integrates previously unseen heterogeneous agents into our CP model by updating only lightweight modules while freezing complex perception components, reducing retraining cost  \textit{and} preserving single-agent performance.

\item \textbf{Scalable real-time deployment.}
In real-time, \textsc{PEARL} is driven by a low-complexity, lightweight model-interpreter selection strategy, which can be parallelized across multiple candidate interpreters and joining agents. Experiments show that PEARL enables practical real-time deployment with only 1.67 ms similarity-scoring latency and efficient large-scale interpreter assignment.

\item \textbf{Real-time interpreter assignment of novel agents.}
\textsc{PEARL} enables \textit{novel agents}, which were not seen during training, to participate in CP sessions in real-time by assigning the most compatible interpreter among the available lightweight interpreters to each joining agent.

\item \textbf{Extensive experiments and ablation studies.} Experiments on OPV2V, V2XSet, and DAIR-V2X datasets show that PEARL’s real-time model selection strategy yields gains \textbf{5.3}\% AP@0.5 and \textbf{8.2}\% AP@0.7 over a random-selection baseline on average. Although designed for real-time CP, PEARL also outperforms state-of-the-art heterogeneous CP under traditional offline training by \textbf{3.3}\% AP@0.5 and \textbf{5.6}\% AP@0.7 for new-agent adaptation and heterogeneous base training, on average, while reducing computation and preserving privacy.

\end{itemize}

\section{Related Work}
\subsection{Collaborative Perception}
Collaborative perception (CP) expands each vehicle’s field-of-view and mitigates occlusions, thereby improving safety and reliability~\cite{huang2023v2x,yazgan2024survey}. CP methods are commonly categorized by the nature of communicated information among collaborating vehicles, and by the corresponding stage at which fusion of shared data takes place within the perception pipelines of these CP vehicles. In \emph{early} fusion, vehicles share raw sensor data, which can yield high accuracy but requires prohibitive communication bandwidth. In \emph{late} fusion, vehicles exchange only final detection results, which reduces bandwidth requirements significantly. However, late CP suffers from the fusion of erroneous detections such as false positives or misaligned detections, and may even lead to higher numbers of mis-detections if not handled properly~\cite{yazgan2024survey,hong2024multi}. Recent CP works mainly use \emph{intermediate} fusion by sharing latent features typically generated by one or more stages of deep-learning backbones. This intermediate CP strategy enables a trade-off between the accuracy of early fusion and the bandwidth efficiency of late fusion.
Aside from the type of CP fusion, there is a large body of work that addresses critical CP challenges. For example, to address transmission-bandwidth constraints, communication-aware frameworks decide what/whom to transmit the shared data to reduce overhead~\cite{hu2024communication,hu2022where2comm,yang2023what2comm}. Other lines of CP work address pose error~\cite{song2024spatial,ni2024self,lei2024robust} and latency~\cite{wang2024intercoop,lei2024robust,zheng2024cooperfuse}. Meanwhile, one of the most critical aspects of CP, which is \emph{heterogeneity} in sensors, models, and training data, remains comparatively less explored than other areas. Below, we briefly review the two most related areas of our work: (a) state-of-the-art methods for heterogeneous CP, and (b) recent efforts in employing visual prompts within learning-based pipelines.

\subsection{Heterogeneous CP}

Addressing domain gaps from heterogeneous sensors, models, and training data is essential for reliable CP. Recent work follows two main strategies: \emph{retraining} models into a unified semantic space or learning \emph{interpreters} that map heterogeneous features into a shared domain. Some works design unified CP architectures that assume known sensor/model configurations and rely on joint retraining across all participating agent types~\cite{li2024di,xiang2023hm,xu2022v2x}, while other methods retrain part of the stack toward standardized semantics~\cite{lu2024extensible,kong2025cobevmoe,shao2025negocollab}. Notably, HEAL~\cite{lu2024extensible} defines a standard feature semantics based on the ego agent's homogeneous CP pipeline, then trains neighboring-agent encoders to match this standard while keeping the standard feature semantics frozen. Alternatively, several approaches learn per-type feature interpreters that align heterogeneous features to an ego-centric space~\cite{xu2022bridging,xin2025pnpda+,xia2025one,lu2025privacy}. While~\cite{xu2022bridging,xin2025pnpda+,lu2025privacy} use two-step interpretation pipelines, PolyInter~\cite{xia2025one} addresses the heterogeneous domain gap with a one-step interpreter that jointly performs channel and spatial alignment. Faster-HEAL~\cite{maleki2026faster} also enables a one-step interpreter using a low-rank visual prompt to address the heterogeneous domain gap. However, these methods typically require either joint retraining, prior knowledge of agent types, or explicit sensor/model metadata to select the appropriate alignment module. Building on these limitations, \textsc{PEARL} proposes an \textbf{anonymous}, \textbf{real-time} interpreter that assigns a model to a newly joining agent \emph{without} disclosing its sensor or model metadata.

\subsection{Visual Prompts}
Visual prompts inject task-relevant information into intermediate features via small learnable tensors and enable models to handle domain shift with frozen backbones and minimal compute. Early work learned \emph{domain-invariant} representations for classification, improving generalization to new domains~\cite{cao2024domain,gan2023decorate,singha2023ad}. Subsequent studies adapted prompts for downstream detection~\cite{farrukh2025prmpt2adpt,zhang2025upre,li2023learning,li2024ada}. Notably,~\cite{yang2024exploring} showed that although dense prompts suit classification, sparse, spatially selective prompts better address detection gaps. Prompt tuning has also been applied to heterogeneous CP via PolyInter~\cite{xia2025one} and Faster-HEAL~\cite{maleki2026faster}, showing prompts provide a compact and efficient mechanism for perception under distribution shift.

\section{Methodology}

\begin{figure*}[t]
  \centering
  \includegraphics[width=0.92\linewidth]{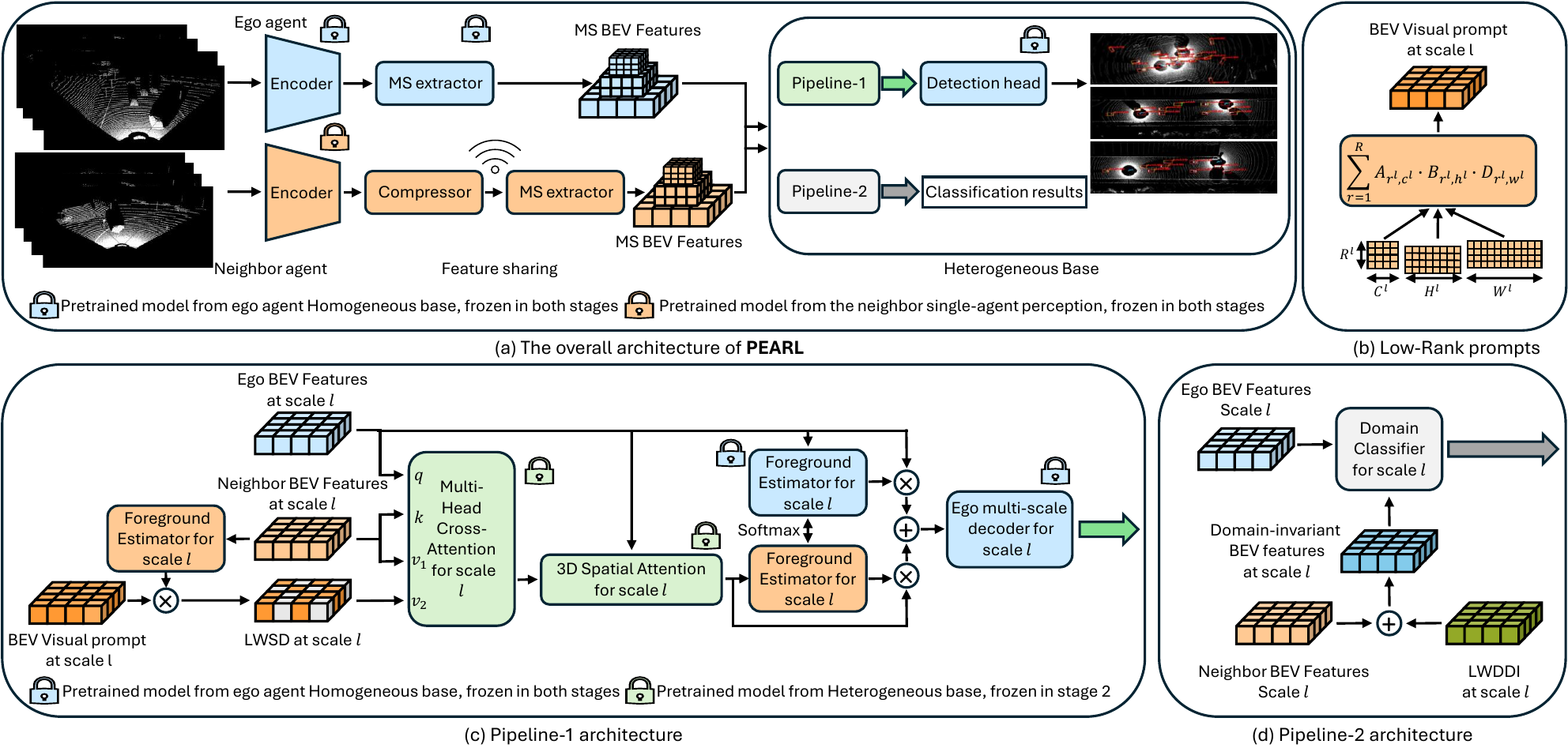}
 \caption{\textbf{(a)} \textsc{PEARL} architecture and offline training. In Stage-1, only neighbor-side modules (compressors, MS extractors, pipelines) are trained, while all encoders and the ego perception stack remain frozen. In Stage-2, only lightweight prompts, the domain classifier, and the new agent’s compressor and MS extractor are fine-tuned.
\textbf{(b)} PARAFAC~\cite{carroll1970analysis}-based prompt reconstruction: three 2D factors over $C,H,W$ with rank $R$ rebuild the 3D prompt and reduce parameters from $C \times H \times W$ to $R(C{+}H{+}W)$ with $R \ll \min(C,H,W)$.
\textbf{(c)} Detection pipeline: ego and compressed neighbor BEV features are channel-aligned by two MHCA branches (neighbor BEVs and $\mathrm{LWSD}$ as values), spatially aligned with 3D attention, fused with foreground-based weights, and upsampled by the ego MS decoder.
\textbf{(d)} Domain-classification pipeline: $\mathrm{LWDDI}$ prompts are added to neighbor BEVs, which with ego BEVs pass through a shared adversarial classifier, yielding domain-invariant features for fast model selection in real-time.}
  \label{fig:four-cvpr}
\end{figure*}
The overall architecture of \textsc{PEARL} is shown in Figure~\ref{fig:four-cvpr}~(a). The framework has two pipelines: (i) a \textbf{detection pipeline} that aligns heterogeneous neighbor features to the ego feature space for cooperative detection, and (ii) an \textbf{adversarial domain-classification pipeline} that learns domain-invariant BEV features. Both pipelines use lightweight visual–prompt interpreters to align features per agent and per scale while preserving privacy. In Stage-1, heterogeneous agents are mapped to the ego semantic space during heterogeneous base training. In Stage-2, we freeze all heavy modules and fine-tune only lightweight prompts and a low-complexity domain classifier for new joining agents, preserving privacy, maintaining single-agent detection quality, and enabling fast adaptation for real-time deployment.
To process heterogeneous inputs, each neighbor's BEV feature tensor is first compressed to match the ego’s channel and spatial dimensions. The ego BEV features and each compressed neighbor feature are then passed through their own multi-scale (MS) extractors to produce scale-aligned, coarse-to-fine BEV features. Both pipelines operate on these multi-scale features: the detection pipeline outputs upsampled features for the detection head, while the domain-classification pipeline uses adversarial training to align MS features to a unified semantic space and obtain domain-invariant features.
\subsection{Pipeline-1: Detection}
\label{ss3:pipe1}
Figure~\ref{fig:four-cvpr}~(c) illustrates the detection pipeline. It aligns multi-scale BEV features from heterogeneous neighbors to the ego feature space and fuses them for cooperative detection. The pipeline takes the ego’s and compressed neighbors' MS BEV features at all scales. Because each scale encodes different semantics and receptive fields, we attach a separate visual prompt to every neighbor type and scale. Inspired by~\cite{xia2025one}, we adapt a channel–selection module and a spatial transformer to further refine BEV features for detection.

We first build sparse prompts focused on detection-relevant regions. At each scale $l$, a foreground estimator $ f_{\text{fg}}^{(l)}$ predicts a softmax occupancy map that gates a learnable prompt:
\begin{equation}
\begin{gathered}
\mathrm{OCC}_{i}^{(l)} = f_{\text{fg}}^{(l)}(F_{i}^{(l)}), \quad i=1,\dots,N, \\
\mathrm{LWSD}_{i}^{(l)} = P_{D,i}^{(l)} \odot \mathrm{OCC}_{i}^{(l)}, \quad i=1,\dots,N,
\end{gathered}
\end{equation}
where $\mathrm{OCC}_{i}^{(l)}$ is the occupancy map for agent type $i$, $F_{i}^{(l)}$ is the BEV feature tensor, $P_{D,i}^{(l)}$ is the randomly initialized prompt for the detection pipeline, and $\mathrm{LWSD}_{i}^{(l)}$ is the resulting sparse detection prompt. We reuse the same foreground estimator for fusion to keep the design lightweight.

Because different encoders and datasets reorder BEV semantic channels, we realign neighbor channels to the ego channels with two multi-head cross-attention (MHCA~\cite{xia2025one}) branches. At scale $l$, we flatten ego features as queries and neighbor features as keys:
\begin{equation}
    Q_{h}^{(l)} = F_{e}^{\prime (l)} W_{q,h}^{(l)}, \quad
    K_{h}^{(l)} = F_{i}^{\prime (l)} W_{k,h}^{(l)}, 
\end{equation}
\begin{equation}
     M_{j,h}^{(l)} = \text{Softmax}\!\left(\frac{Q_{h}^{(l)} {K_{h}^{(l)}}^\top}{\sqrt{d_k^{(l)}}}\right), \quad j=1,2
\end{equation}
where $F^{\prime (l)}$ are the flattened features, $h$ indexes attention heads, $j$ the MHCA branch, $d_k^{(l)}$ the key dimension, and $W_{q,h}^{(l)}$, $W_{k,h}^{(l)}$ learned projection matrices. We use two value streams,
\begin{equation}
    V_{1,h}^{(l)} = F_{i}^{\prime (l)} W_{v_1,h}^{(l)}, \qquad
    V_{2,h}^{(l)} = \mathrm{LWSD}_{i}^{\prime (l)} W_{v_2,h}^{(l)},
\end{equation}
with $W_{v_1,h}^{(l)}$, $W_{v_2,h}^{(l)}$ as value projections. The first branch captures feature-to-feature similarity, and the second injects prompt-guided alignment. We then fuse the outputs:
\begin{equation}
\begin{gathered}
    F^{(l)}_{j,h} = M_{j,h}^{(l)} V_{j,h}^{(l)}, \quad
    F^{(l)}_{j} = \text{concat}_h F^{(l)}_{j,h}, \\
    F^{(l)}_{\text{ca}} = \text{LN}(F^{(l)}_{1}) + \text{LN}(F^{(l)}_{2}),
\end{gathered}
\end{equation}
where $\mathrm{LN}(\cdot)$ is layer normalization, yielding channel-aligned BEV features $F^{(l)}_{\text{ca}}$ that emphasize detection-relevant regions via $\mathrm{LWSD}$.

We also need spatial alignment for effective fusion after channel-wise alignment. We apply a 3D spatial attention module~\cite{xu2022cobevt} between $F_{e}^{(l)}$ and $F^{(l)}_{\text{ca}}$ at scale $l$; since $F^{(l)}_{\text{ca}}$ highlights detection-relevant regions, the attention module more accurately aligns neighbor features to the ego features.

Finally, we perform multi-scale fusion. For each agent and scale, the foreground estimator predicts a spatial objectness probability; we normalize it per agent and use it as fusion weight~\cite{lu2024extensible}. This process is lightweight (only a weighted sum) and reuses the same estimator as $\mathrm{LWSD}$, so weights match the prompted regions. After fusion, features at each scale are upsampled by the ego multi-scale decoder to a common resolution, concatenated, and fed to the detection head.
\subsection{Pipeline-2: Adversarial Domain Classification}
\label{ss3:pipe2}
Figure~\ref{fig:four-cvpr}~(d) illustrates the second pipeline. Its goal is to make neighbor BEV features domain-invariant to the ego, so they are comparable across heterogeneous vehicles and enable fast model selection in real-time. As in pipeline-1, it receives the ego’s multi-scale BEV features and the compressed neighbor BEV features. For each scale and neighbor type, we attach a dense visual prompt to align that scale’s features to the ego space before classification:
\begin{equation}
     F_{DI,i}^{(l)} = F_{i}^{(l)} + \mathrm{LWDDI}_{i}^{(l)},
\end{equation}
\begin{equation}
    D_{i}^{(l)} = f_{\text{cls}}(F_{DI,i}^{(l)}), \quad
    D_{e}^{(l)} = f_{\text{cls}}(F_{e}^{(l)}),
\end{equation}
where $\mathrm{LWDDI}_{i}^{(l)}$ is a lightweight prompt for agent type $i$ at scale $l$ in the adversarial domain-classification pipeline, $F_{i}^{(l)}$ and $F_{e}^{(l)}$ are neighbor and ego BEV features, and $f_{\text{cls}}$ is a shared domain classifier.

We train $\mathrm{LWDDI}$ and the classifier adversarially: the classifier minimizes domain-classification error, while the prompts make domains indistinguishable,
\begin{equation}
\label{eq:adv-loss-short}
    \mathcal{L}_{\text{adv},i}^{(l)} =
    \max_{\text{LWDDI}} \min_{D} \;
    \mathcal{L}_{\text{BCE}}(D_{e}^{(l)}) + \mathcal{L}_{\text{BCE}}(D_{i}^{(l)}),
\end{equation}
where $\mathcal{L}_{\text{adv},i}^{(l)}$ is the loss for agent type $i$ at scale $l$, $D_{e}^{(l)}$ and $D_{i}^{(l)}$ are ego and neighbor classifier outputs, and $\mathcal{L}_{\text{BCE}}$ is binary cross-entropy. We realize the min–max optimization with a gradient-reversal layer~\cite{ganin2015unsupervised}. Since this pipeline uses only lightweight prompts and a shallow classifier, it is fast, which we will take advantage of for real-time deployment.
\subsection{Low-Rank Visual Prompts}
\label{ss3:visualprompt}
Both pipelines use visual prompts represented as 3D tensors of size $C_l \times H_l \times W_l$ at scale $l$. With two prompts per scale and agent ($\mathrm{LWSD}$ and $\mathrm{LWDDI}$), the total parameter count is \[
2  N \sum _l  C_l  H_l  W_l,
\]
which quickly becomes impractical for high-resolution BEVs or many agents. Typically, downstream perception tasks do not require such a large amount of additional visual information, leading to slow convergence and overfitting.

We therefore construct the prompt with a low-rank PARAFAC (CP) factorization~\cite{carroll1970analysis}: instead of learning the whole tensor, we learn three 2D factors for channel, height, and width with decomposition rank $R$,
\begin{equation}
    A \in \mathbb{R}^{R \times C}, \quad
    B \in \mathbb{R}^{R \times H}, \quad
    D \in \mathbb{R}^{R \times W},
\end{equation}
and reconstruct as shown in Figure~\ref{fig:four-cvpr}~(b),
\begin{equation}
    \label{equ:cp}
    P_{c,h,w} \approx \sum_{r=1}^{R} A_{r,c} \, B_{r,h} \, D_{r,w},
\end{equation}
where $P_{c,h,w}$ is the visual-prompt value at channel $c$ and spatial location $(h, w)$. The PARAFAC factorization reduces the number of parameters per prompt to $R(C+H+W)$ with $R \ll \min(C,H,W)$, reducing computation overhead while maintaining high accuracy for both pipelines during training, and making Stage-2 and real-time implementation feasible.
\subsection{Stage-1: Heterogeneous Base Training}
Stage-1 builds a heterogeneous base so that (i) detection remains accurate under BEV domain gaps and (ii) a new agent can be adapted with only lightweight updates. As shown in Figure~\ref{fig:four-cvpr}, we jointly train both pipelines together with neighbor-side compressors, multi-scale extractors, and foreground estimators,  while treating the ego BEV feature space as the reference, and all heterogeneous agents are mapped to this unified space. To preserve privacy and improve efficiency, neighbors send only intermediate BEV features (no model or sensor metadata), while all encoders and the ego detection stack (encoder, multi-scale extractor/decoder, foreground estimator, detection head) remain frozen.

Detection-only supervision is insufficient to close heterogeneous gaps, so we add a semantic alignment loss and a single-agent detection loss for neighbors~\cite{xia2025one}. For each neighbor $i$ and scale $l$, we match the mean and variance of the cross-attended neighbor feature to the ego feature:
\begin{equation}
\label{eq:det-loss}
\begin{gathered}
    \mathcal{L}^{(l)}_{\text{dis1},i}
    = \big\| \mu (F^{\prime (l)}_{ca}) - \mu (F^{(l)}_e)\big\|_2
    + \big\| \sigma (F^{\prime (l)}_{ca}) - \sigma (F^{(l)}_e)\big\|_2, \\
    \mathcal{L}_{\text{det},i}
    = \mathcal{L}_{\text{box},i} + \mathcal{L}_{\text{reg},i},
\end{gathered}
\end{equation}
where $F^{\prime (l)}_{ca}$ and $F^{(l)}_e$ are the flattened cross-attention outputs for neighbor $i$ and ego at scale $l$, $\mu(\cdot)$ and $\sigma(\cdot)$ are mean and standard deviation, and $|\cdot|_2$ is the $L_2$ distance. Here $\mathcal{L}^{(l)}_{\text{dis1},i}$ measures the semantic distance between ego and neighbor in pipeline-1, $\mathcal{L}_{\text{box},i}$ and $\mathcal{L}_{\text{reg},i}$ are box classification and regression losses, and $\mathcal{L}_{\text{det},i}$ is the single-neighbor detection loss. The total pipeline-1 loss is
\begin{equation}
    \mathcal{L}_{\text{pipe1}}
    = \mathcal{L}_{\text{collab}}
    +  \alpha_{1} \sum_{i=1}^{N} \mathcal{L}_{\text{det},i}+ \beta_{\text{dis1}} \sum_{l=1}^{L} \sum_{i=1}^{N}
         \mathcal{L}^{(l)}_{\text{dis1},i} ,
\end{equation}
where $\alpha_{1}$ and $\beta_{\text{dis1}}$ are loss-balancing weights. $\mathcal{L}_{\text{collab}}$ is the collaborative detection loss on the fused, upsampled multi-scale features fed to the ego detection head, enforcing correctness of the final cooperative output. The per-agent $\mathcal{L}_{\text{det},i}$ terms act as auxiliary single-neighbor supervision that speeds training and stabilizes domain alignment.

In parallel, pipeline-2 learns domain-invariant features. In addition to the adversarial loss in~(\ref{eq:adv-loss-short}), we match moments between prompted neighbor and ego features:
\begin{equation}
\label{eq:det-loss-prime}
    \mathcal{L}^{(l)}_{\text{dis2},i}
    = \big\| \mu ( F^{(l)}_{DI,i} ) - \mu ( F^{(l)}_{e} ) \big\|_2
    + \big\| \sigma ( F^{(l)}_{DI,i} ) - \sigma ( F^{(l)}_{e} ) \big\|_2,
\end{equation}
where $F^{(l)}_{DI,i}$ and $F^{(l)}_{e}$ are domain-invariant features for neighbor $i$ and the ego at scale $l$. The total loss for pipeline-2 is
\begin{equation}
    \mathcal{L}_{\text{pipe2}}
    = \sum_{l=1}^{L} \sum_{i=1}^{N}
      \big( \alpha_{2} \mathcal{L}^{(l)}_{\text{adv},i}
          + \beta_{\text{dis2}} \mathcal{L}^{(l)}_{\text{dis2},i} \big).
\end{equation}
where $\alpha_{2}$ and $\beta_{\text{dis2}}$ are loss-balancing hyperparameters. Stage-1 is trained end-to-end with
\begin{equation}
    \mathcal{L}_{\text{stage1}} = \mathcal{L}_{\text{pipe1}} + \mathcal{L}_{\text{pipe2}}.
\end{equation}
\subsection{Stage-2: New Heterogeneous Agent Training}
The Stage-1 base enables fast adaptation when a new agent joins. We unfreeze only $\mathrm{LWSD}$ and the foreground estimator in pipeline-1, $\mathrm{LWDDI}$ and the domain classifier in pipeline-2, and the new agent's multi-scale extractor and compressor; all other components remain frozen. This maps the new agent’s BEV features into the established heterogeneous base space, while preserving efficiency and privacy, since the ego still receives only BEV features. We build new agent-type interpreters on top of the shared heterogeneous base (MHCA and spatial attention), so the ego can host many interpreters while storing only one base module.
\subsection{Real-Time Implementation}
\label{ss3:real}
There are several underexplored challenges in real-time heterogeneous CP. To the best of our knowledge, no single model or interpreter can handle the full diversity of agent types the ego may encounter at runtime. In practice, the ego must rely on offline training to build multiple models or interpreters, each tailored to one agent type, and then select a suitable one for each newly joining agent at deployment.

In this context, real-time CP faces three main challenges: (i) most schemes require sensor and model metadata from the new agent to choose an interpreter, which violates privacy, (ii) “seen” agent types in offline training can produce different BEV semantics when their encoders change due to different training data, reintroducing domain gaps, (iii) and completely unseen agent types may appear, for which offline retraining is not immediately possible. Yet even unseen agents often share partial semantics (e.g., the same backbone with a different configuration), so the ego should select the closest pretrained interpreter based solely on BEV features. Naïve strategies, e.g., assigning a random model can degrade performance significantly (as we show later), or exhaustively retraining/testing pretrained models in real-time, are simply not feasible (even then, there are no ground truth labels for accurate training).

To address these issues, \textsc{PEARL} uses pipeline-2’s domain-invariant features to select a suitable pretrained interpreter in real-time using only BEV features. Figure~\ref{fig:real-time-selection} illustrates the one-time anonymous model-selection process and the subsequent collaborative detection stage. The ego first provides a shape-compatible compressor so the new agent’s BEV features match its expected size and reduce bandwidth. After receiving the compressed BEVs, the ego runs these features through each interpreter's multi-scale extractor with the corresponding $\mathrm{LWDDI}$ prompt to obtain domain-invariant features:
\begin{equation}
    F^{(l)}_{DI,i} = g^{(l)}_i( F_{n}) + \mathrm{LWDDI}_{i}^{(l)}, \quad i = 1,\dots,N,
\end{equation}
where $g^{(l)}_i$ is scale $l$ of candidate interpreter i’s extractor, $F_{n}$ is the new agent’s BEV. The ego then compares spatially averaged, L2-normalized features $\bar{F}_{DI,i}$ to its own $\bar{F_e}$ using cosine similarity:

\begin{equation}
\begin{gathered}
    S_{i} = 
    \sum_{l=1}^{L} w^{(l)}
    \left\langle
        \frac{\bar{F}^{(l)}_{DI,i}}{\|\bar{F}^{(l)}_{DI,i}\|_2},
        \frac{\bar{F}^{(l)}_{e}}{\|\bar{F}^{(l)}_{e}\|_2}
    \right\rangle,
    \quad i^{*} = \arg\max_{i} S_{i},
\end{gathered}
\end{equation}
The interpreter with index $i^{*}$ is selected for the new agent.

\begin{figure*}[t]
  \centering
  \includegraphics[width=0.76\linewidth]{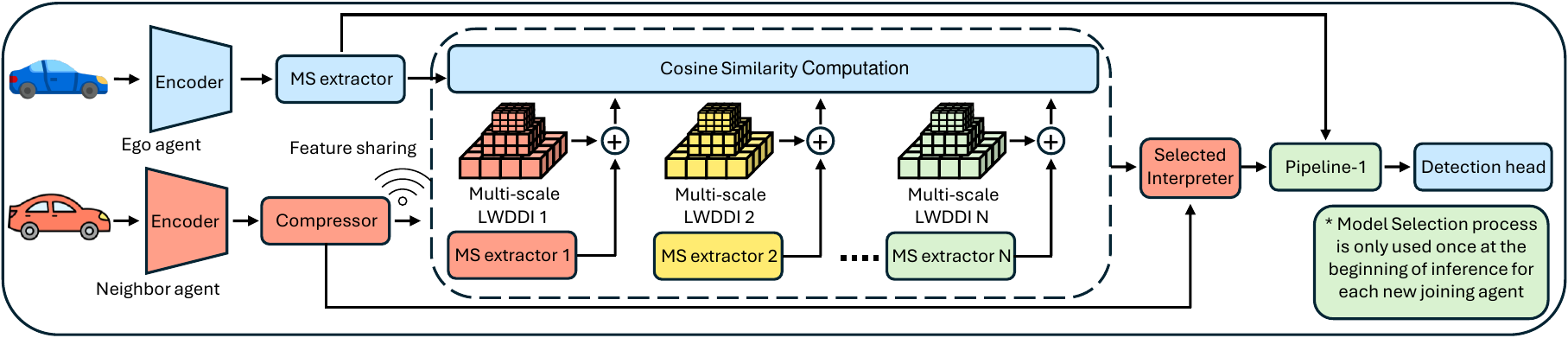}
 \caption{Real-time anonymous interpreter selection and deployment in \textsc{PEARL}. When a new heterogeneous agent joins, the ego first provides a shape-compatible compressor, and the joining agent transmits only compressed BEV features without revealing its sensor type, model architecture, or training configuration. The ego then evaluates the joining agent's features through the candidate multi-scale \textsc{LWDDI} paths and computes cosine similarity between the resulting domain-invariant features and the ego features. The interpreter with the highest similarity is selected as the most compatible candidate. This model-selection process is performed only once when the agent joins. After selection, the chosen interpreter is used in Pipeline-1 for subsequent collaborative detection, where the aligned neighbor features are fused with the ego features and passed to the detection head.}
\label{fig:real-time-selection}

\label{fig:real-time-selection}
\end{figure*}

Because both Stages already reduced semantic distance in both pipelines [Eqs.~(\ref{eq:det-loss}) and (\ref{eq:det-loss-prime})], the similarity score correlates with detection performance. The selection step uses only lightweight matrix–vector operations, runs in \textbf{1.67 ms}, and preserves privacy since only BEV features are shared. A single stored model can serve multiple agents whose similarity exceeds a threshold. Together with Stage-2’s lightweight adaptation, this fast selection spawns many interpreters from one heterogeneous base and ego backbone with negligible storage, enabling real-time operation for diverse agent types.

\section{Experimental Results}
\label{Sec:4}
\subsection{Dataset and Experimental Setting}
\label{sec:exp}

We evaluate \textsc{PEARL} on three large-scale collaborative perception datasets, OPV2V~\cite{xu2022opv2v}, V2XSet~\cite{xu2022v2x}, and DAIR-V2X~\cite{yu2022dair}. Unless otherwise stated, we report the ablations on OPV2V, while V2XSet and DAIR-V2X are used to validate generalization under different data distributions, with DAIR-V2X further reflecting real-world cooperative perception scenarios.
On DAIR-V2X, each collaborative sample contains one infrastructure ego agent and one vehicle neighbor, allowing us to evaluate \textsc{PEARL} under a real-world vehicle-to-infrastructure collaborative-perception setting.
\paragraph{Heterogeneous setting}
Following prior work on heterogeneous collaborative perception, we simulate realistic multi-agent scenarios with diverse sensor encoders and domain gaps. We adopt three LiDAR-based encoder families: PointPillars (pp)~\cite{lang2019pointpillars}, SECOND (sd)~\cite{yan2018second}, and VoxelNet (vn)~\cite{zhou2018voxelnet}. 
The ego agent uses a pp8 encoder, where the suffix denotes voxel-size variant, producing a BEV feature map of size $64 \times 64 \times 128$. Neighboring agents use heterogeneous encoders, including pp4, sd2 and vn4 in Stage-1, and sd1, vn6, or pp4 in Stage-2. The encoder specifications are shown in~Appendix~\ref{sec:encoder}. All encoder outputs are passed through lightweight compressors to match the ego feature size.

We adopt a two-stage training protocol. In Stage-1 (heterogeneous base), the model is trained with an ego agent (pp8) and two heterogeneous neighbors (vn4 with either pp4 or sd2), exposing cross-domain feature discrepancies while keeping training efficient. In Stage-2 (new-agent adaptation), we introduce one unseen agent at a time (sd1, vn6, or pp4) and fine-tune only a subset of modules, including the prompts, compressor, multi-scale (MS) extractor, foreground estimator, and domain classifier. The encoders, ego MS extractor and decoder, detection head, and other Pipeline-1 components remain frozen. This setting evaluates the ability of \textsc{PEARL} to generalize to unseen agents with minimal adaptation.

For multi-modal experiments, we additionally incorporate image-based encoders using ResNet~\cite{he2016deep} and EfficientNet~\cite{tan2019efficientnet}, forming heterogeneous configurations such as pp8-vn4-ResNet and pp8-pp4-EfficientNet in Stage-1, and pp8-ResNet or pp8-EfficientNet in Stage-2.

\paragraph{Architecture details}
\textsc{PEARL} operates on a compact ego feature space to ensure efficiency. The ego BEV representation has $64$ channels and spatial resolution $64 \times 128$, and all modules in Pipeline-1 and Pipeline-2 are designed around this feature size. Neighbor features are compressed to this space to reduce communication and accelerate convergence.

The multi-scale (MS) extractor consists of three ResNeXt~\cite{xie2017aggregated} stages. The first preserves feature resolution with $64$ channels, while the next two progressively downsample feature maps' spatial resolution by a factor of $2$ while doubling the number of channels, producing outputs of size $128 \times 32 \times 64$ and $256 \times 16 \times 32$, respectively. All alignment modules, including channel cross-attention, spatial attention, foreground estimation, \textsc{LWSD}, \textsc{LWDDI}, and the domain classifier, operate across these three scales. The MS decoder upsamples each scale to a common resolution and aggregates them via channel concatenation before detection.

The MS extractor and decoder are initialized from the pretrained ego backbone, and the ego extractor is kept frozen during training. The low-rank prompt decomposition uses ranks $R=[8,16,32]$ across scales, enabling efficient adaptation while preserving expressiveness.

\paragraph{Training details}
We use a batch size of 8 and train all models on a single NVIDIA RTX A6000 GPU. The loss weights are set to $\beta_{\text{dis1}} = \beta_{\text{dis2}} = 0.5$, $\alpha_1 = 1$, and $\alpha_2 = 0.05 \times 9.1$. Additional optimization and evaluation details are provided in Appendix~\ref{app:training_details}.

\subsection{Performance Comparison}
\label{ss4:perf}
We compare \textsc{PEARL} with the state-of-the-art PolyInter~\cite{xia2025one} under the authors’ settings, using \textsc{CoBEVT}~\cite{xu2022cobevt} for PolyInter’s fusion and matching LiDAR ranges for fairness. Tables~\ref{tab:stage1_both}-\ref{tab:stage2_lidar_other} summarize results on OPV2V, V2XSet, and DAIR-V2X for Stage-1 (heterogeneous base) and Stage-2 (new-agent adaptation). \textsc{PEARL} consistently reduces heterogeneous domain gaps in Stage-1 and further adapts interpreters in Stage-2, while maintaining real-time capability (see Section~\ref{ss3:real}).

\paragraph{Stage-1: Heterogeneous base}
As shown in Table~\ref{tab:stage1_both}, with pp8 as ego and LiDAR neighbors (pp4-vn4, vn4-sd2), \textsc{PEARL} improves over PolyInter by \textbf{+2.4/+4.5\%} and \textbf{+2.3/+5.5\%} on OPV2V, and by \textbf{+2.3/5.0\%} and \textbf{+2.4/5.1\%} on V2XSet, for AP@0.5/@0.7 respectively. 
On DAIR-V2X, \textsc{PEARL} also achieves the largest gains of \textbf{+6.2/+8.2\%} and \textbf{+5.3/+7.8\%} under the same settings.
These gains indicate more effective alignment of heterogeneous feature distributions across datasets, and suggest that \textsc{PEARL} transfers particularly well to real-world data.
For cross-modality settings on OPV2V (vn4-ResNet, pp4-EfficientNet), \textsc{PEARL} further improves performance by up to \textbf{+2.2/+4.0\%}, demonstrating robustness to modality gaps.

\begin{table*}[t]
\centering
\caption{Detection performance comparison (AP@0.5 / AP@0.7) with \textsc{CoBEVT} fusion for PolyInter on OPV2V, V2XSet, and DAIR-V2X under Stage-1 scenarios.}
\label{tab:stage1_both}
\resizebox{0.85\linewidth}{!}{%
\begin{tabular}{c|cccc|cc|cc}
\toprule
Dataset
& \multicolumn{4}{c|}{OPV2V}
& \multicolumn{2}{c|}{V2XSet}
& \multicolumn{2}{c}{DAIR-V2X} \\
\cmidrule(lr){2-5}\cmidrule(lr){6-7}\cmidrule(lr){8-9}
Stage-1 scenario
& pp4-vn4 & vn4-sd2 & vn4-ResNet & pp4-EfficientNet
& pp4-vn4 & vn4-sd2
& pp4-vn4 & vn4-sd2 \\
\midrule
\textbf{PEARL (ours)}
& \textbf{95.5/83.6}
& \textbf{96.3/85.4}
& \textbf{92.3/77.7}
& \textbf{92.5/78.3}
& \textbf{91.3/71.7}
& \textbf{91.2/71.8}
& \textbf{72.1/45.3}
& \textbf{72.4/45.2} \\
PolyInter~\cite{xia2025one}
& 93.1/79.1
& 94.0/79.9
& 90.1/73.7
& 91.4/74.5
& 89.0/66.7
& 88.8/66.7
& 65.9/37.1
& 67.1/37.4 \\
\bottomrule
\end{tabular}
}
\end{table*}

\paragraph{Stage-2: OPV2V LiDAR and camera-agent adaptation}
Table~\ref{tab:stage2_opv2v} reports interpreter adaptation to new agents on OPV2V, including both LiDAR-agent (pp4, sd1, vn6) and camera-agent (EfficientNet, ResNet) settings. 
For LiDAR-agent adaptation, \textsc{PEARL} outperforms PolyInter by \textbf{+1.4/+2.0\%} (pp4), \textbf{+0.9/+1.3\%} (sd1), and up to \textbf{+4.0/+6.2\%} (vn6) for AP@0.5/@0.7 respectively. 
The largest improvements occur on vn6, where longer LiDAR range and coarser voxelization introduce larger domain shifts, highlighting the effectiveness of our adaptation strategy. 
For camera-agent adaptation, \textsc{PEARL} improves over PolyInter by \textbf{+1.7/+2.2\%} for ResNet and \textbf{+3.1/+3.3\%} for EfficientNet. 
Notably, gains are more pronounced at AP@0.7, indicating better localization under modality discrepancies.

\begin{table*}[t]
\centering
\caption{Detection performance (AP@0.5 / AP@0.7) under Stage-2 new-agent adaptation on OPV2V, including LiDAR-agent (pp4, sd1, vn6) and camera-agent (EfficientNet, ResNet) settings.}
\label{tab:stage2_opv2v}
\resizebox{0.85\linewidth}{!}{%
\begin{tabular}{c|ccc|ccc|c|c}
\toprule
Stage-1 scenario
& \multicolumn{3}{c|}{pp4-vn4}
& \multicolumn{3}{c|}{vn4-sd2}
& vn4-ResNet
& pp4-EfficientNet \\
\cmidrule(lr){2-4}\cmidrule(lr){5-7}\cmidrule(lr){8-8}\cmidrule(lr){9-9}
Stage-2 scenario
& pp4 & sd1 & vn6
& pp4 & sd1 & vn6
& EfficientNet
& ResNet \\
\midrule
\textbf{PEARL (ours)}
& \textbf{95.5/83.7}
& \textbf{95.7/85.1}
& \textbf{90.7/73.7}
& \textbf{95.6/83.9}
& \textbf{95.8/86.5}
& \textbf{90.7/73.7}
& \textbf{89.8/73.1}
& \textbf{89.5/72.8} \\
PolyInter~\cite{xia2025one}
& 93.9/81.9
& 94.6/84.5
& 86.7/67.5
& 94.4/81.8
& 95.1/84.6
& 87.2/68.5
& 86.7/69.8
& 87.8/70.6
 \\
\bottomrule
\end{tabular}
}
\end{table*}

\paragraph{Stage-2: V2XSet and DAIR-V2X LiDAR-agent adaptation}
As shown in Table~\ref{tab:stage2_lidar_other}, similar trends are observed on V2XSet, with gains of +1.6/+6.7\% (pp4), +1.6/+6.0\% (sd1), and +5.3/+8.3\% (vn6) under the pp4-vn4 setting, and +2.5/+8.7\% (pp4), +1.5/+5.7\% (sd1), and +3.9/+7.7\% (vn6) under vn4-sd2, for AP@0.5/@0.7 respectively. On DAIR-V2X, \textsc{PEARL} further demonstrates strong adaptation capability, achieving gains of \textbf{+5.8/+7.7\%} (pp4), \textbf{+6.2/+7.7\%} (sd1), and \textbf{+6.4/+7.6\%} (vn6) under the pp4-vn4 setting, and \textbf{+5.3/+8.4\%} (pp4), \textbf{+5.5/+8.1\%} (sd1), and \textbf{+5.7/+8.3\%} (vn6) under vn4-sd2, for AP@0.5/@0.7 respectively. These results indicate that \textsc{PEARL} generalizes effectively to real-world cooperative perception scenarios under both heterogeneous initialization and new-agent adaptation.

\begin{table*}[t]
\centering
\caption{Detection performance (AP@0.5 / AP@0.7) under Stage-2 LiDAR-agent adaptation (pp4, sd1, vn6) on V2XSet and DAIR-V2X.}
\label{tab:stage2_lidar_other}
\resizebox{0.9\linewidth}{!}{%
\begin{tabular}{c|cccccc|cccccc}
\toprule
Dataset
& \multicolumn{6}{c|}{V2XSet}
& \multicolumn{6}{c}{DAIR-V2X} \\
\cmidrule(lr){2-7}\cmidrule(lr){8-13}
Stage-1 scenario
& \multicolumn{3}{c|}{pp4-vn4}
& \multicolumn{3}{c|}{vn4-sd2}
& \multicolumn{3}{c|}{pp4-vn4}
& \multicolumn{3}{c}{vn4-sd2} \\
\cmidrule(lr){2-4}\cmidrule(lr){5-7}\cmidrule(lr){8-10}\cmidrule(lr){11-13}
Stage-2 scenario
& pp4 & sd1 & vn6
& pp4 & sd1 & vn6
& pp4 & sd1 & vn6
& pp4 & sd1 & vn6 \\
\midrule
\textbf{PEARL (ours)}
& \textbf{91.2/71.7}
& \textbf{91.3/73.0}
& \textbf{84.6/59.2}
& \textbf{91.4/72.3}
& \textbf{91.6/72.8}
& \textbf{83.7/59.5}
& \textbf{71.6/45.1}
& \textbf{72.8/45.5}
& \textbf{68.2/42.7}
& \textbf{71.6/44.8}
& \textbf{72.7/45.5}
& \textbf{68.5/43.2} \\
PolyInter~\cite{xia2025one}
& 89.6/65.0
& 89.7/67.0
& 79.3/50.9
& 88.9/63.6
& 90.1/67.1
& 79.8/51.8
& 65.8/37.4 & 66.6/37.8 & 61.8/35.1
& 66.3/36.4 & 67.2/37.4 & 62.8/34.9 \\
\bottomrule
\end{tabular}
}
\end{table*}

\subsection{Model efficiency and communication cost}
\label{Sec:trainable}
We further analyze the efficiency of \textsc{PEARL} in terms of trainable parameters and communication overhead, both of which are critical for real-world deployment.

\paragraph{Trainable parameters}
Compared to PolyInter~\cite{xia2025one}, \textsc{PEARL} uses fewer trainable parameters during Stage-1, reducing the parameter count by \textbf{1.1M--4.4M} across heterogeneous bases. In Stage-2, \textsc{PEARL} adapts only a lightweight set of modules (prompts, compressor, MS extractor, and domain classifier), resulting in a compact adaptation model of approximately \textbf{4.1M} parameters. Although PolyInter may use fewer parameters in some Stage-2 cases, its cumulative trainable-parameter requirement across the two training stages remains higher due to its heavier Stage-1 adaptation design. Overall, \textsc{PEARL} achieves a lower total parameter count in most scenarios while delivering superior detection performance, owing to its lightweight prompt-based formulation. Trainable parameters comparison is shown in~Appendix~\ref{sec:params}.

\paragraph{Communication efficiency}
\textsc{PEARL} significantly reduces communication overhead by transmitting only compressed intermediate BEV features of size $64 \times 64 \times 128$ ($\sim$0.52M elements). As shown in Table~\ref{Tab:size}, this is substantially smaller than the feature representations used in PolyInter across different encoder backbones. Depending on the configuration, \textsc{PEARL} reduces the transmitted feature size by \textbf{$8.7\times$ to $34.7\times$}, with an average reduction of approximately \textbf{$24\times$}. Since communication cost scales linearly with the number of transmitted elements under fixed precision, this reduction directly translates to lower bandwidth requirements.

\begin{table}[t]
\centering
\caption{Comparison of transmitted intermediate BEV feature sizes between \textsc{PEARL} and PolyInter across encoder backbones. Feature size is reported as tensor dimensions.}
\label{Tab:size}
\setlength{\tabcolsep}{5pt}
\resizebox{0.86\columnwidth}{!}{%
\begin{tabular}{l|c|c|c}
\toprule
\textbf{Method / Encoder} & \textbf{Feature Shape} & \textbf{\# Elements} & \textbf{Reduction} \\
\midrule
\textbf{PEARL (ours)} & $64 \times 64 \times 128$ & $\sim$0.52M & 1$\times$ \\
\midrule
PolyInter (pp4) & $384 \times 100 \times 352$ & $\sim$13.52M & $\sim$26.0$\times$ \\
PolyInter (vn6) & $128 \times 128 \times 512$ & $\sim$8.39M & $\sim$16.1$\times$ \\
PolyInter (vn4) & $128 \times 200 \times 704$ & $\sim$18.02M & $\sim$34.7$\times$ \\
PolyInter (sd2) & $512 \times 50 \times 176$ & $\sim$4.51M & $\sim$8.7$\times$ \\
PolyInter (sd1) & $512 \times 100 \times 352$ & $\sim$18.02M & $\sim$34.7$\times$ \\
\bottomrule
\end{tabular}
}
\end{table}

\subsection{Real-Time Implementation}
\label{ss4:real}

We evaluate \textsc{PEARL} under a realistic real-time deployment setting, where the ego vehicle must assign an appropriate interpreter to a newly joining heterogeneous agent using only the received intermediate BEV features. This setting is motivated by three practical challenges. First, the joining agent may not disclose its sensor or model metadata, making conventional metadata-based assignment undesirable from both privacy and deployment perspectives. Second, even when the joining agent belongs to a nominally known type, differences in training data, initialization, or learned weights can induce semantic mismatch in the intermediate BEV features, which degrades collaborative performance if the wrong interpreter is selected. Third, completely unseen agent types may appear at runtime, for which retraining or online validation is not feasible. In such cases, rejecting the agent or assigning an interpreter at random can significantly reduce detection accuracy, while exhaustive online evaluation of all candidate models is too costly and would require unavailable ground-truth annotations.

\begin{figure*}
    \centering
    \includegraphics[width=0.93\linewidth]{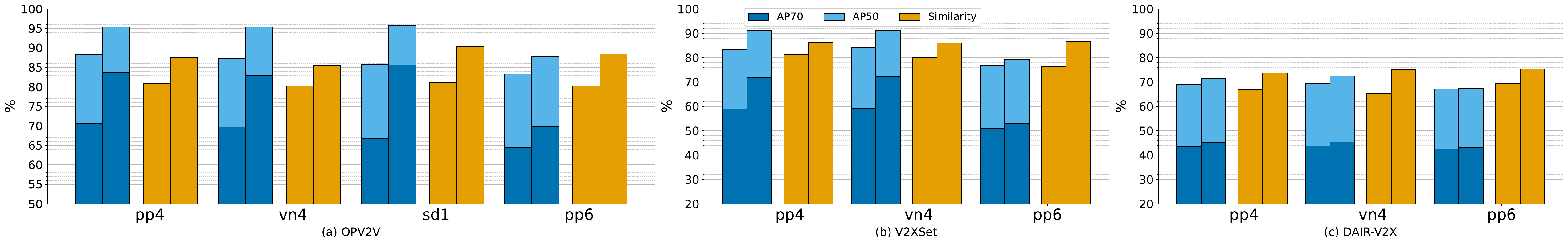}
    \caption{\textbf{Summary of real-time anonymous interpreter selection on OPV2V, V2XSet, and DAIR-V2X.}
    For each joining-agent type, the left bar reports the average detection accuracy over all candidate interpreters, corresponding to uniform random selection. For known joining-agent types—pp4, vn4, and sd1—the right bar reports the average performance of interpreters pretrained on the corresponding agent type. For the previously unseen pp6 agent, the right bar reports the performance of the highest-similarity selected interpreter. Gold bars show the corresponding cosine similarity in the domain-invariant space. Across the three datasets, higher similarity is consistently associated with stronger collaborative detection performance.}
    \label{fig:real-time}
\end{figure*}

To address these issues, \textsc{PEARL} performs interpreter assignment in the domain-invariant feature space produced by Pipeline-2, as illustrated in Figure~\ref{fig:real-time-selection}. After receiving the compressed BEV features from the joining agent, the ego applies each pretrained $\mathrm{LWDDI}$ path and computes the cosine similarity between the resulting domain-invariant features and its own features. The interpreter associated with the highest similarity is then selected for collaborative detection. We use three feature scales, with stride~2 between consecutive scales and channel width doubling across scales, and aggregate the scale-wise cosine similarities using weights $[1,2,4]$. Importantly, this assignment is carried out entirely in the domain-invariant space because domain-specific features from the other agents are not accessible at deployment; obtaining them would require running their encoders, exposing sensor/model details, and introducing substantial computational overhead. In contrast, the proposed similarity-based assignment requires only compressed BEV features and lightweight feature processing on the ego side, making it well-suited for privacy-preserving real-time collaboration.
To quantify the effectiveness of this strategy, we compare the selected interpreter against the average performance across all available pretrained interpreters, which serves as a random-selection baseline in the absence of an established benchmark for anonymous interpreter assignment. For known joining-agent types, we additionally report the average performance across interpreters pretrained on the corresponding agent type. This evaluates whether the learned similarity consistently retrieves compatible interpreters across different heterogeneous bases, rather than depending on a single specific interpreter. Figure~\ref{fig:real-time} summarizes the results on (a) OPV2V, (b) V2XSet, and (c) DAIR-V2X.
On OPV2V, we evaluate four anonymous joining agents, namely pp4, vn4, sd1, and pp6. In all cases, the interpreters pretrained on the matching agent type achieve both higher feature similarity and stronger collaborative detection performance than the average over all candidate interpreters. For pp4, the matching interpreters improve performance by +7.0\% AP@0.5 and +13.0\% AP@0.7 over the average across all interpreters, while their mean similarity is higher by about 0.06. For vn4, the corresponding gains are +8.1\% AP@0.5 and +13.3\% AP@0.7, with an average similarity margin of approximately 0.05. For sd1, the gains further increase to +10.0\% AP@0.5 and +18.9\% AP@0.7, with a similarity advantage of about 0.09. For the unseen pp6 agent, the matching top-selected interpreter improves performance by +4.5\% AP@0.5 and +5.5\% AP@0.7, with a similarity advantage of about 0.08.  These results show a consistent alignment between cosine similarity in the learned domain-invariant space and downstream collaborative detection quality, indicating that the proposed assignment criterion is not only efficient but also highly predictive of the best-performing interpreter. 

We also evaluate the same anonymous assignment protocol on V2XSet. The trends remain consistent with those observed on OPV2V, despite the different dataset characteristics. For an anonymous pp4 joining agent, the interpreters pretrained on pp4 outperform the average over all candidate interpreters by +8.0\% AP@0.5 and +12.8\% AP@0.7, while achieving a similarity advantage of about 0.05. Likewise, when vn4 acts as the anonymous joining agent, the interpreters pretrained on vn4 improve over the average by +7.1\% AP@0.5 and +12.9\% AP@0.7, with a similarity margin of about 0.06. For the unseen pp6 agent, the selected top interpreter improves over the average by +2.5\% AP@0.5 and +2.1\% AP@0.7, while increasing similarity by about 0.10. Although the absolute AP values differ from OPV2V, the same qualitative pattern holds: the candidate interpreter with the highest similarity in the domain-invariant space is also the one that yields the best collaborative performance. 

On DAIR-V2X, the same behavior is observed under a real-world deployment benchmark. For an anonymous pp4 joining agent, the interpreters pretrained on pp4 exceed the average baseline by +2.8\% AP@0.5 and +1.5\% AP@0.7, while achieving a similarity advantage of about 0.07. For an anonymous vn4 joining agent, interpreters pretrained on vn4 improve over the average by +2.9\% AP@0.5 and +1.7\% AP@0.7, with a similarity margin of about 0.10. For the unseen pp6 agent, the selected top interpreter slightly improves detection performance by +0.3\% AP@0.5 and +0.6\% AP@0.7, while still increasing similarity by about 0.06. These results further confirm that higher similarity in the learned domain-invariant space remains a reliable indicator of better collaborative performance, even under real-world sensing conditions. 

Beyond accuracy, runtime efficiency and scalability are essential for practical deployment. We therefore measure both the lightweight similarity-scoring step in Pipeline-2 and the full interpreter selection process. On average, the similarity computation alone takes only \textbf{1.67 ms} per sample, as it operates on domain-invariant features using cosine similarity. When the cost of the multi-scale (MS) feature extractor is included, the average end-to-end interpreter selection time becomes \textbf{48.67 ms} per sample. For a representative \texttt{pp4} interpreter adapted on the \texttt{pp4-vn4} base, the similarity-scoring step takes \textbf{1.6 ms}, and the MS extractor takes \textbf{27.7 ms}, yielding \textbf{29.3 ms} end-to-end. Importantly, this overhead is incurred only once when a new agent joins; after assignment, the ego uses only the selected interpreter for subsequent collaborative detection. As a result, interpreter selection remains significantly cheaper than full collaborative inference. 

\begin{figure}
    \centering
    \includegraphics[width=0.87\linewidth]{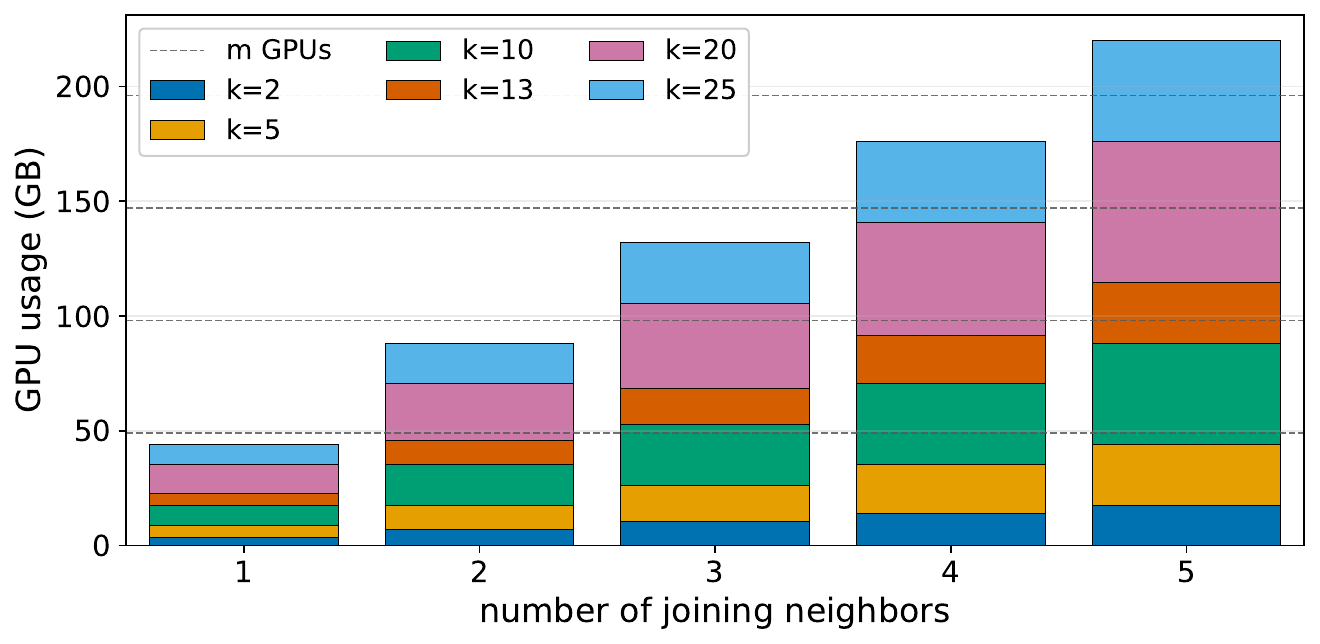}
\caption{Scalability of real-time interpreter selection. GPU memory usage as a function of the number of candidate interpreters $k$ and joining agents $n$. Memory grows approximately linearly with $k$ and $n$, but interpreters can be evaluated in parallel within GPU limits. This enables near-constant selection latency (per agent) despite increasing candidate pool size.}
\label{fig:GPU}
\end{figure}

While the above measurements reflect the cost per interpreter, in practice, \textsc{PEARL} evaluates multiple candidate interpreters in parallel. The selection process is fully decoupled from the detection pipeline and can be executed concurrently across interpreters on modern GPUs. As a result, the effective latency does not scale linearly with the number of candidate interpreters, but is instead bounded by hardware parallelism and memory availability.
To analyze this behavior, we study the memory footprint required to evaluate multiple interpreters simultaneously. As shown in Figure~\ref{fig:GPU}, the GPU memory usage grows approximately linearly with both the number of candidate interpreters $k$ and the number of joining agents $n$. For a large candidate pool ($k=25$), the total memory requirement is approximately $44$~GB, which fits within a single modern GPU (e.g., RTX A6000). This enables all interpreters to be evaluated concurrently, resulting in an effective selection latency comparable to a single-interpreter evaluation (i.e., $\approx 48.67$ ms), rather than scaling with $k$. 

Furthermore, when multiple agents join simultaneously, the selection process can be parallelized across agents, with total memory scaling with $k \times n$. Practical configurations such as $(k=13, n=2)$ can be handled on a single GPU, while larger settings can be distributed across multiple GPUs with near-linear throughput scaling. 

Overall, these results demonstrate that \textsc{PEARL} achieves a favorable trade-off between accuracy, privacy, and runtime: interpreter assignment is fast, scalable, and incurs only a one-time overhead, making real-time deployment feasible even in dynamic and heterogeneous multi-agent environments. Detailed results are reported in Appendix~\ref{sec:real-appendix}. 

\subsection{Cross-Dataset Tests}

To evaluate robustness under dataset-level domain shift, cross-dataset model-selection experiments are conducted using the \texttt{pp4} encoder trained on OPV2V. The corresponding real-time selection results for OPV2V-trained interpreters are provided in Appendix~\ref{sec:cross-dataset-detailed}. Table~\ref{Tab:pp4_cross} summarizes performance when selecting among interpreters trained on OPV2V versus V2XSet. AP@0.5/0.7 and the average cosine similarity score $S_{\mathrm{avg}}$ are reported, computed over the top-3 interpreters with the highest similarity in each case.

Interpreters trained on the same dataset as the joining agent, namely OPV2V, achieve higher detection performance, reaching 95.4/83.7\% AP@0.5/0.7 with $S_{\mathrm{avg}}=0.8746$. In comparison, the top-3 interpreters trained on V2XSet achieve 87.7/60.4\% AP@0.5/0.7 with $S_{\mathrm{avg}}=0.8420$. This gap reflects the impact of dataset-level domain shift on intermediate BEV representations, even when the joining agent uses the same nominal encoder type, \texttt{pp4}. These results demonstrate that sensor or model metadata alone is insufficient to determine interpreter compatibility under distribution shift. Instead, \textsc{PEARL}'s feature-based similarity selection provides a principled mechanism to identify compatible interpreters at runtime while adapting to both intra- and cross-dataset variations.
\begin{table}[t]
\centering
\caption{Cross-dataset evaluation for the pp4 encoder trained on OPV2V. We compare the model selection performance of the top-3 selected interpreters for candidate interpreters trained on the OPV2V against the V2XSet dataset.} 

\label{Tab:pp4_cross}
\setlength{\tabcolsep}{6pt}
\renewcommand{\arraystretch}{1.05}
\resizebox{0.95\columnwidth}{!}{%
\begin{tabular}{l|c|c}
\toprule
Top-3 Interpreters trained on &  \textbf{OPV2V}  & V2XSet \\
\midrule
AP@0.5/0.7 (\%)  &\textbf{95.4/83.7} & 87.7/60.4 \\
$S_{\mathrm{avg}}$ & \textbf{0.8746} & 0.8420 \\
\bottomrule
\end{tabular}%
}
\end{table}

\subsection{Model Selection Baseline}
To evaluate the effectiveness of \textsc{PEARL}’s model selection strategy, comparisons are performed against several baseline approaches using the same set of pretrained interpreters. While a random-selection baseline establishes a lower bound under anonymous settings, it does not isolate the contribution of individual design components. Two additional baselines are therefore introduced to analyze the impact of (i) domain-invariant feature learning and (ii) the similarity metric:
\textbf{Baseline 1 (B1):} selection using cosine similarity computed directly on intermediate features, without \textsc{LWDDI}.
\textbf{Baseline 2 (B2):} selection using \textsc{LWDDI}, but replacing cosine similarity with a moment-matching distance.

Table~\ref{Tab:pp4_baselines} reports the average performance over the top-3 selected interpreters for the \texttt{pp4} encoder. \textsc{PEARL} consistently outperforms both baselines, achieving improvements of +3.8/+4.8\% in AP@0.5 and +6.9/+8.4\% in AP@0.7 over B1 and B2, respectively. These results demonstrate that both components of \textsc{PEARL}’s selection mechanism, i.e., domain-invariant representations (\textsc{LWDDI}) and cosine similarity matching, are critical for robust and accurate interpreter selection under heterogeneous and anonymous settings.

\begin{table}[t]
\centering
\caption{Comparison of \textsc{PEARL} with baseline selection strategies for the pp4 encoder. B1 performs selection using cosine similarity without domain-invariant features (\textsc{LWDDI}). B2 uses \textsc{LWDDI} with a moment-matching distance instead of cosine similarity. Results report the average over the top-3 highest similarities.}
\label{Tab:pp4_baselines}
\setlength{\tabcolsep}{6pt}
\renewcommand{\arraystretch}{1.05}
\resizebox{0.95\columnwidth}{!}{%
\begin{tabular}{l|c|c|c}
\toprule
Top-3 Interpreters average &  \textbf{PEARL}  & B1 & B2\\
\midrule
AP@0.5/0.7 (\%)  &\textbf{95.4/83.7} & 91.6/76.8 &90.6/75.3\\
\bottomrule
\end{tabular}%
}
\end{table}

\subsection{Ablation Study}
\paragraph{Component-wise evaluation (pipeline-1)}
\label{ss4:abl}
We evaluate the contribution of the main components and design choices in Pipeline-1, including the foreground estimator for $\mathrm{LWSD}$, multi-head cross-attention, 3D spatial attention, the multi-scale extractor design, scale-specific $\mathrm{LWSD}$ prompts, and foreground-weighted fusion. Specifically, we compare \textsc{PEARL} against the following variants: \texttt{-MS extractor}, which replaces the multi-scale design with a single-scale model; \texttt{Shared extractor}, which uses the same frozen extractor weights for all agents; \texttt{-FG for LWSD}, which removes the foreground estimator from sparse prompt generation; \texttt{-Cross-attention}, which removes the channel-alignment module; \texttt{-Spatial attention}, which removes the spatial alignment module; \texttt{Single \textsc{LWSD}}, which uses one prompt before the MS extractor instead of scale-specific prompts; and \texttt{No fusion weights}, which replaces learned foreground-based fusion with a fixed weight of 0.5. For all settings, we train Stage-1 with \texttt{pp8} as ego and \texttt{vn4} together with \texttt{pp4} as neighbors, and then independently adapt new agents in Stage-2 (\texttt{pp4}, \texttt{sd1}, and \texttt{vn6}).

The results in Table~\ref{tab:ablmodel1} show that every component contributes to the final performance. The largest degradation is caused by removing the MS extractor, which reduces AP by 9.3/19.3\% at AP@0.5/AP@0.7 in Stage-1 and by 10.8/19.2\% on average in Stage-2. This confirms that coarse-to-fine multi-scale processing is important for heterogeneous feature alignment, particularly for challenging cases like \texttt{vn6} with a larger domain gap. The next strongest effect comes from removing cross-attention, which lowers AP by 5.8\% and 2.2\% at AP@0.5, and by 3.5\% and 8.0\% at AP@0.7 in Stage-1 and Stage-2, indicating the importance of channel-wise semantic alignment across heterogeneous encoders.

Using a shared frozen extractor also causes a consistent drop, reducing AP by 1.0/3.7\% in Stage-1 and 1.6/4.7\% in Stage-2 on average, which suggests that heterogeneous agents require distinct coarse-to-fine transformations even after feature compression. Removing spatial attention leads to a smaller but still consistent degradation of 1.6/3.9\% in Stage-1 and 1.0/2.5\% in Stage-2, confirming that spatial alignment remains necessary after channel alignment. Likewise, removing the foreground estimator from $\mathrm{LWSD}$ causes a modest drop, showing that sparse prompting is more effective when guided by detection-relevant regions.

The additional design ablations further support the proposed formulation. Using a single $\mathrm{LWSD}$ prompt across all scales reduces performance by 3.6/8.2\% in Stage-1 and by 2.6/6.8\% in Stage-2 on average, demonstrating that prompt specialization across scales is important for capturing heterogeneous semantics at different resolutions. Replacing learned foreground-weighted fusion with 0.5 also degrades performance by 1.1/1.3\% in Stage-1 and 1.4/1.5\% in Stage-2 on average, indicating that adaptive spatial fusion better exploits the complementary information provided by neighboring agents. 
\begin{table}[t]
\centering
\caption{Effect of pipeline-1 components and extractor design. -MS extractor: single-scale model. Shared extractor: same frozen extractor weights for all agents. Single \textsc{LWSD}: using a single \textsc{LWSD} instead of separate \textsc{LWSD} prompts for each scale. No fusion weights: using 0.5 as fusion weights.}
\label{tab:ablmodel1}
\resizebox{0.93\columnwidth}{!}{%
    \begin{tabular}{c|cccc}
    \toprule
         Model Variant & pp4-vn4 & pp4& sd1 & vn6 \\
        \midrule
        \textbf{PEARL} &\textbf{95.5/83.6}&\textbf{95.5/83.7}&\textbf{95.7/85.1}&\textbf{90.7/73.7}\\
        -MS extractor & 86.2/64.3 & 86.1/65.6&87.0/69.9&76.3/49.3\\
        Shared extractor&94.5/79.9&94.7/80.1&93.9/78.3&88.4/70.1\\
        -FG for LWSD & 95.1/83.1&95.2/83.4&95.4/84.6&89.9/73.2\\
        -Cross-attention & 89.7/80.1 & 92.2/74.1& 93.3/74.9& 89.8/73.5\\
        -Spatial attention & 93.9/79.7 &  94.3/80.5&94.7/82.8&90.0/71.7 \\
         Single \textsc{LWSD} & 91.9/75.4&92.7/74.6&93.4/78.7&87.9/68.9\\
        No fusion weights & 94.4/82.3&94.6/82.5&94.7/83.8&88.4/71.6\\
        \bottomrule
    \end{tabular}
    }
\end{table}

\paragraph{Effect of PARAFAC rank}
We evaluated the PARAFAC (CP) rank $R$ for prompt construction. Larger $R$ values inject more capacity but can slow convergence or overfit; smaller $R$ values may underfit alignment. As shown in Table~\ref{tab:ablmode2}, $R{=}[8,16,32]$ offers the best trade-off with $\approx30$K parameters. Across settings, PARAFAC prompts use $\approx12$–$50$K weights ($<0.5\%$ of total Stage-1) and have comparable performance, versus $\sim1.83$M for dense 3D prompts, which slows training and undermines generalization, and has lower performance.
\begin{table}[t]
\centering
\caption{Impact of PARAFAC rank on AP (Stage-1 pp4-vn4; Stage-2 new-agent adaptation). Prompt parameters range from 12K–50K with PARAFAC vs. $\sim$1.83M for dense 3D prompts.}
\resizebox{0.93\columnwidth}{!}{%
    \begin{tabular}{c|cccc}
    \toprule
       PARAFAC Rank & pp4-vn4 & pp4& sd1 & vn6 \\
        \midrule
        \texttt{3D prompt} &95.4/83.2&95.3/83.1&95.6/85.1&90.5/73.0\\
        
         \texttt{[8,8,8]} &95.3/83.4&95.2/83.2&95.5/84.6&90.4/73.3   \\        \textbf{\texttt{[8,16,32]}}&9\textbf{5.5/83.6}&\textbf{95.5/83.7}&\textbf{95.7/85.1}&\textbf{90.7/73.7}\\
        \texttt{[16,16,16]}&95.5/83.1&95.4/83.6&95.5/85.1&90.5/73.5 \\
        \texttt{[32,32,32]}&95.4/83.3&95.3/83.6&95.4/85.0&90.3/73.2 \\
        \bottomrule
    \end{tabular}
    }
    
    \label{tab:ablmode2}
\end{table}

\section{Conclusion}
We presented \textbf{PEARL}, an anonymous and real-time framework for heterogeneous collaborative perception that employs visual prompts as interpreters to mitigate the domain gap introduced by heterogeneous sensors and models. PEARL is designed for real-time deployment, where a new anonymous agent can join without disclosing sensor or model metadata and where offline retraining is impractical. \textsc{PEARL} combines two lightweight interpreters, $\mathrm{LWDDI}$ for generating domain-invariant features, and $\mathrm{LWSD}$ for detection-critical alignment. During real-time operation, an unseen or anonymous agent can be assigned an appropriate pretrained model through a fast, privacy-preserving process by comparing domain-invariant features from $\mathrm{LWDDI}$ and selecting the most similar model, whose corresponding $\mathrm{LWSD}$ is then applied for cooperative detection. \textsc{PEARL} also supports efficient offline adaptation for new agent types with minimal computation while preserving privacy, making it a practical and scalable solution for heterogeneous collaborative perception under real-world constraints.

\section{Acknowledgment}
This work was supported in part by the US Department of Transportation under grant 69A36523420190CRSCA.
\bibliographystyle{IEEEtran}
\bibliography{main}
\vspace{-10pt}

\section{Biography}
\vspace{-35pt}

\begin{IEEEbiography}[{
\includegraphics[width=1in]{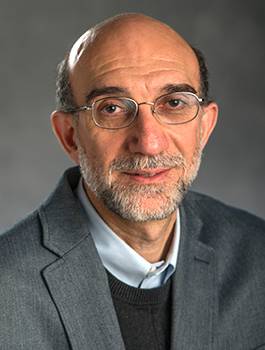}}]{Hayder Radha} \textit{(Fellow, IEEE)} received the Ph.M. and Ph.D. degrees from Columbia University, New York, NY, USA, in 1991 and 1993, respectively. He was a Fellow and Principal Member of Research Staff at Philips Research from 1996 to 2000 and a Distinguished Member of Technical Staff and MTS at Bell Laboratories from 1986 to 1996. He is currently a MSU Foundation Distinguished Professor and the Director of the Connected and Autonomous Networked-Vehicles for Active Safety (CANVAS) Program, Michigan State University. He is a recipient of the Sony Research Award, the Amazon Research Award, the Semiconductor Research Consortium Award, two Google Faculty Research Awards, two Microsoft Research Awards, the AT\&T Bell Labs Ambassador, AT\&T Circle-of-Excellence Awards, and the William J. Beal Outstanding Faculty Award.
\end{IEEEbiography}
\vspace{-30pt}
\begin{IEEEbiography}[{
\includegraphics[width=1in]{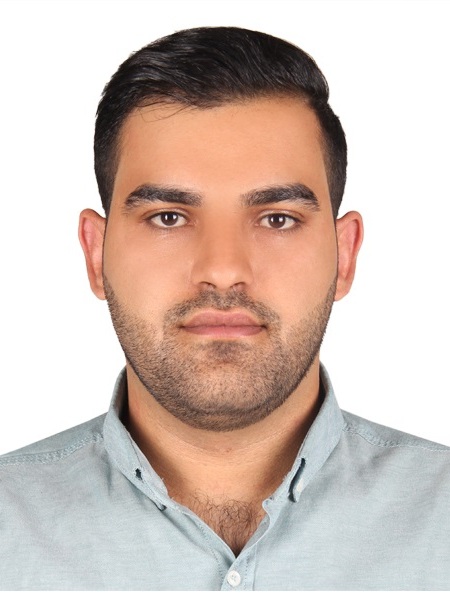}}]{Armin Maleki}
\textit{(Graduate Student Member, IEEE)} received the M.S. degree in electrical engineering from the University of Notre Dame, Notre Dame, IN, USA, in 2024. He is currently pursuing the Ph.D. degree in electrical and computer engineering at Michigan State University, East Lansing, MI, USA. His research interests include collaborative perception, autonomous vehicles, computer vision, signal processing, and multi-agent data fusion. He is a member of the IEEE Computer Society and the IEEE Intelligent Transportation Systems Society.
\end{IEEEbiography}

\clearpage

\appendices
\section{LiDAR Encoders Details}
\label{sec:encoder}
Table~\ref{Tab1} summarizes the LiDAR encoder configurations, including the voxel resolution, the 3D CNN layers that process the encoder input, and the 2D BEV backbone that converts 3D features into a BEV feature map with the sizes shown in the last column. Unless otherwise specified, all encoders operate over a LiDAR range of $[-102.4, 102.4]\,\mathrm{m}$ along $x$ and $[-51.2, 51.2]\,\mathrm{m}$ along $y$. The pp6 and vn6 variants instead use an extended range of $[-153.6, 153.6]\,\mathrm{m}$ along $x$ and $[-76.8, 76.8]\,\mathrm{m}$ along $y$. For PointPillar~\cite{lang2019pointpillars}, we use three variants. The pp8 variant serves as the ego encoder and produces a BEV feature map of size $64 \times 64 \times 128$. The pp4 variant is used as a heterogeneous neighbor in both Stage-1 and Stage-2. For VoxelNet~\cite{zhou2018voxelnet}, we use vn4 as a heterogeneous neighbor in Stage-1 and vn6 as a new heterogeneous agent in Stage-2 for adaptation. Finally, for SECOND~\cite{yan2018second}, we use sd2 as a heterogeneous neighbor in Stage-1 and sd1 as a new heterogeneous neighbor in the adaptation stage. All encoder variations are followed by a lightweight compressor that maps their BEV features to the ego feature size of $64 \times 64 \times 128$.

\begin{table}[t]
  \centering
    \caption{Detailed configurations for agent type encoders.}
  \setlength{\tabcolsep}{3pt}
  \renewcommand{\arraystretch}{1.05}
  \resizebox{\columnwidth}{!}{
  \begin{tabular}{c| c| c | c| c |c}
    \toprule
    Encoder & Variation &
    Voxel &
    2D / 3D &
    Half LiDAR &
    Feature size \\
    &&Resolution&CNN Layers&Range (x,y)&($C\times H\times W$)\\
    \midrule
    \multirow{3}{*}{PointPillar~\cite{lang2019pointpillars}}
      &  pp8 & 0.8, 0.8, 4   & 4 / 0   & 102.4, 51.2   & $64\times64\times128$\\
      &  pp6 & 0.6, 0.6, 4   & 4 / 0   & 153.6, 76.8 & $64\times128\times256$ \\
      &  pp4 & 0.4, 0.4, 4   & 4 / 0   & 102.4, 51.2   & $64\times128\times256$ \\
    \midrule
    \multirow{2}{*}{VoxelNet~\cite{zhou2018voxelnet}}
      &  vn6 & 0.6, 0.6, 0.4 & 0 / 3    & 153.6, 76.8 & $128\times256\times512$ \\
      &  vn4 & 0.4, 0.4, 0.4 & 0 / 3    & 102.4, 51.2   & $128\times256\times512$ \\
    \midrule
    \multirow{2}{*}{SECOND~\cite{yan2018second}}
      &  sd2 & 0.2, 0.2, 0.2 & 4 / 12  & 102.4, 51.2   & $64\times64\times128$ \\
      &  sd1 & 0.1, 0.1, 0.1 & 4 / 13  & 102.4, 51.2   & $64\times128\times256$ \\
    \bottomrule
  \end{tabular}}
  \label{Tab1}
\end{table}

\section{Additional Training and Evaluation Details}
\label{app:training_details}

All models are implemented in PyTorch and trained on a single NVIDIA RTX A6000 GPU with a batch size of 8. Table~\ref{tab:training_details} summarizes the optimization settings used in the two training stages. In Stage~1, multi-head cross-attention, 3D spatial attention, the neighboring-agent compressors, multi-scale extractors, foreground estimators, LWSD and LWDDI prompts, and domain classifiers are optimized, while all agent encoders and the ego perception stack remain frozen. In Stage~2, training is initialized from the corresponding Stage~1 checkpoint, and only the new agent's compressor and multi-scale extractor, the foreground estimator, LWSD and LWDDI prompts, and domain classifier are updated. All remaining modules are frozen.

\begin{table}[t]
\centering
\caption{Optimization settings used for PEARL training.}
\label{tab:training_details}
\setlength{\tabcolsep}{4pt}
\begin{tabular}{lc}
\toprule
Setting & Both training stages \\
\midrule
Optimizer & Adam \\
Initial learning rate & 0.001 \\
Weight decay & 0.0001 \\
Number of epochs & 35 \\
Learning-rate scheduler & MultiStepLR \\
Scheduler milestones & [10,25] \\
Learning-rate decay factor & 0.1 \\
Gradient clipping & None \\
Numerical precision & FP32  \\
\bottomrule
\end{tabular}
\end{table}

Unless otherwise specified, the loss weights are $\alpha_1=1$, $\beta_{\mathrm{dis1}}=\beta_{\mathrm{dis2}}=0.5$, and $\alpha_2=0.05\times9.1$. The gradient-reversal scaling follows the adversarial domain-alignment formulation in~\cite{ganin2015unsupervised}. Model checkpoints are selected according to the highest validation AP@0.5 metric. The reported results are obtained from 28 separately trained experimental configurations.

We use the official training, validation, and test splits of OPV2V, V2XSet, and DAIR-V2X. The corresponding split sizes are 6374/1980/2170, 6694/609/2834, and 4811/1789/1789, respectively. 
For DAIR-V2X, the infrastructure agent is treated as the ego agent and the vehicle as the heterogeneous neighboring agent. Detection performance is evaluated using AP at IoU thresholds of 0.5 and 0.7, following the official evaluation protocol of each dataset.

Runtime is measured with a batch size of 1. Before timing, we perform 50 warm-up iterations, followed by 10 measured iterations. CUDA synchronization is applied immediately before and after each timed operation, and the reported latency is averaged over all measured samples. GPU memory is reported as the peak allocated memory during inference.

\section{Comparison of Trainable Parameter Counts}
\label{sec:params}
We further compare the number of trainable parameters of \textsc{PEARL} and PolyInter~\cite{xia2025one} in offline training. As shown in Table~\ref{Tab:NTC}, in Stage-1 \textsc{PEARL} consistently uses fewer trainable parameters than PolyInter, reducing the parameter count by $1.1$ M--$4.4$ M across different heterogeneous bases. In Stage-2, \textsc{PEARL} has a comparable number of trainable parameters to PolyInter for pp4 and sd1, and a higher number of parameters for the other cases. Overall, and by considering both stages, \textsc{PEARL} has fewer trainable parameters for all scenarios (except for the EfficientNet case) while achieving better detection performance as reported in Section~\ref{Sec:trainable}, and  \textsc{PEARL} is more efficient and more viable for real deployment. The trainable parameter count reflects the optimization, adaptation, and model-update storage burden, and \textsc{PEARL} maintains a more compact overall model than PolyInter due to its lightweight prompt-based design. 
\begin{table*}[t]
\centering
\caption{Number of trainable parameters comparison in million (M) with \textsc{CoBEVT} fusion for PolyInter. Stage-1: heterogeneous bases pp4-vn4, vn4-sd2, vn4-ResNet, pp4-EfficientNet with pp8 as ego. Stage-2: new-agent adaptation per base (LiDAR-only: pp4, sd1, vn6; cross-modality: EfficientNet, ResNet). For each method, the first row corresponds to Stage-1 and the second row to Stage-2.}

\label{Tab:NTC}
 \resizebox{0.9\linewidth}{!}{%
\begin{tabular}{c|ccc|ccc|c|c}
\toprule
Stage-1 Scenarios & &pp4-vn4& & &vn4-sd2 & & vn4-ResNet &  pp4-EfficientNet \\
\cmidrule(lr){1-1} \cmidrule(lr){3-3} \cmidrule(lr){6-6}\cmidrule(lr){8-8}  \cmidrule(lr){9-9} 
 Stage-2 Scenarios & pp4 & sd1& vn6&  pp4 & sd1& vn6& EfficientNet& ResNet \\
\midrule
\textbf{PEARL (ours)}   & & 12.4 & & &12.4 & & 12.4 & 12.4\\
 \cmidrule(lr){3-3} \cmidrule(lr){6-6}\cmidrule(lr){8-8}  \cmidrule(lr){9-9} 
&4.1&4.1&4.1&4.1&4.1&4.1&4.1&4.1\\
\midrule
PolyInter~\cite{xia2025one} & &15.7& & &16.8 & &13.5&15.7 \\
 \cmidrule(lr){3-3} \cmidrule(lr){6-6}\cmidrule(lr){8-8}  \cmidrule(lr){9-9} 
&4.3&3.2&1.1&4.3&3.2&1.1&1.1&1.1\\
\bottomrule
\end{tabular}
}
\end{table*}

\section{Detailed Real-Time Implementation Results}
\label{sec:real-appendix}
We evaluated \textsc{PEARL} under real-time constraints, as described in~\ref{ss4:real}. At deployment, the ego applies each pretrained \textsc{LWDDI} to the new agent’s intermediate BEV features, computes the cosine similarity to its own domain-invariant features, and assigns the interpreter with the highest similarity, without requiring any sensor or model metadata. We operate in the domain-invariant space because domain-specific features from other agents are not available at deployment; producing them would require access to their encoders, reveal sensor/model details, and incur substantial extra computation. We use three scales (stride~2 per scale, with channels doubling across scales) and weight similarities by $[1,2,4]$ when averaging across scales.

For fairness, we report both detection performance and cosine similarities across all pretrained interpreters (from Stage-1 and Stage-2), to quantify the effectiveness of the selection process. As there is no established baseline for anonymous interpreter assignment, we compare against the average detection AP obtained by randomly selecting an interpreter. We consider a pp4 agent that joins without exposing metadata.

\subsection{OPV2V}
\label{sec:real-opv2v}
A summary of the real-time implementation results is provided in Section~\ref{ss4:real} and Figure~\ref{fig:real-time}. Here, we present the detailed results for pp4 in Table~\ref{Tab:pp4real}. For the pp4 modality, an anonymous agent initiates collaboration with the ego agent. First, the ego shares a compressor weight with the new agent, specified only by the BEV feature size. The new agent then sends only the compressed BEV features back to the ego. This design reduces communication bandwidth while preserving the new agent’s privacy and anonymity. The ego then selects a suitable interpreter by computing the similarity between the new agent’s domain-invariant features, generated by each pretrained \textsc{LWDDI}, and its own domain-invariant features.

For pp4, the detailed results are shown in Table~\ref{Tab:pp4real}. Interpreters that were trained with pp4 (shown in bold) achieve the highest average similarities ($S_{\mathrm{avg}}$) and the best detection performance. Interpreters pretrained on pp4 gain up to \textbf{+24.5\%} AP@0.5 and \textbf{+31.2\%} AP@0.7. On average, the same pretrained interpreters on pp4 gain \textbf{+7.0\%} AP@0.5 and \textbf{+13.0\%} AP@0.7 over the average performance across all interpreters (i.e., under random selection). The mean similarity between the pp4 domain-invariant features and the ego features for interpreters pretrained on pp4 exceeds the all-interpreter average by $0.06$ (with similarity threshold $\tau{=}0.85$).

\begin{table*}[t]
\centering
\caption{Detection performance (AP@0.5 / AP@0.7(\%)), feature similarity for scales $0$--$2$ ($S_0$--$S_2$) and their average ($S_{\mathrm{avg}}$) for real-time anonymous model selection with pp4. Heterogeneous bases: pp4-vn4, vn4-sd2, vn4-ResNet, pp4-EfficientNet with pp8 as ego. Stage-2 (s-2): new-agent adaptation per base (LiDAR-only: pp4, sd1, vn6; cross-modality: EfficientNet, ResNet).}

\label{Tab:pp4real}
 \resizebox{0.80\linewidth}{!}{%
\begin{tabular}{c|c|c|ccc|c}
\toprule
Stage-1 Scenarios & Interpreter &AP@0.5/0.7(\%)& $S_0$&$S_1$ & $S_2$&$S_{avg}$\\
\midrule
\textbf{pp4-vn4} &\textbf{ pp4} & \textbf{95.6/84.0}& \textbf{0.8049} & \textbf{0.8543} &\textbf{0.8649}&\textbf{0.8533} \\
pp4-vn4 &vn4& 93.4/72.9 &0.8844&0.8464&0.8507&0.8543\\
\textbf{pp4-vn4} &\textbf{pp4 (s-2)} &\textbf{95.5/83.7}&\textbf{0.9240}&\textbf{0.8472}&\textbf{0.8913}&\textbf{0.8834}\\
pp4-vn4 &sd1 (s-2)&87.2/66.6&0.8569&0.8149&0.7377&0.7767\\
pp4-vn4 &vn6 (s-2) &70.9/52.5&0.8001&0.7835&0.8184&0.8059 \\
\midrule
vn4-sd2 & vn4 & 84.1/63.6&0.8299&0.7795&0.8327&0.8171\\
vn4-sd2 &sd2 &86.4/63.0&0.8951&0.8058&0.8167&0.8248\\
\textbf{vn4-sd2} &\textbf{pp4 (s-2)} &\textbf{95.5/84.0}&\textbf{0.8403}&\textbf{0.8577}&\textbf{0.86676}&\textbf{0.8604} \\
vn4-sd2 &sd1 (s-2)&88.5/67.1&0.8571&0.8585&0.7653&0.8050\\
vn4-sd2 &vn6 (s-2)&79.7/55.6& 0.8001&0.6654&0.6579&0.6805\\
\midrule
vn4-ResNet & vn4 &88.1/68.6&0.8833&0.8223&0.8376&0.8398\\
vn4-ResNet &ResNet &88.8/71.7&0.8833&0.8400&0.6085&0.7139\\
vn4-ResNet &EfficientNet (s-2)&88.6/71.9&0.9006&0.7671&0.6786&0.7356 \\
\midrule
\textbf{pp4-EfficientNet} & \textbf{pp4} &\textbf{95.2/83.1}&\textbf{0.8428}&\textbf{0.9066}&\textbf{0.9134}&\textbf{0.9014}\\
pp4-EfficientNet & EfficientNet &88.3/71.4&0.8856&0.8733&0.7007&0.7764 \\
pp4-EfficientNet &ResNet (s-2)& 88.4/70.5&0.8191&0.8150&0.7920&0.8025\\
\midrule
\midrule
Average over &  all& 88.4/70.7&0.8568&0.8210&0.7896&0.8082\\
Minimum AP &all&70.9/52.5&0.8001&0.7835&0.8184&0.8059\\
\textbf{Average over}  &\textbf{pretrained pp4} &\textbf{95.4/83.7}&\textbf{0.8530}&\textbf{0.8664}&\textbf{0.8841}&\textbf{0.8746}\\
\bottomrule
\end{tabular}
}
\end{table*}

In addition to pp4, we also use vn4 and sd1 as anonymous newly joining agents to further test the effectiveness of our interpreter selection strategy. 
The detailed results for vn4 are shown in Table~\ref{Tab:vn4real}. Interpreters that were trained with vn4 (shown in bold) achieve the highest average similarities ($S_{\mathrm{avg}}$) and the best detection performance. Interpreters pretrained on vn4 gain up to \textbf{+16.5\%} AP@0.5 and \textbf{+23.6\%} AP@0.7, and \textbf{+8.1\%} AP@0.5 and \textbf{+13.3\%} AP@0.7 over the average performance across all interpreters (i.e., under random selection). The mean similarity between the vn4 domain-invariant features and the ego features for interpreters pretrained on vn4 exceeds the average over all interpreters by $0.05$ (with similarity threshold $\tau{=}0.83$).

\begin{table*}[t]
\centering

\caption{Detection performance (AP@0.5 / AP@0.7(\%)), feature similarity for scales $0$--$2$ ($S_0$--$S_2$) and their average ($S_{\mathrm{avg}}$) for real-time anonymous model selection with vn4. Heterogeneous bases: pp4-vn4, vn4-sd2, vn4-ResNet, pp4-EfficientNet with pp8 as ego. Stage-2 (s-2): new-agent adaptation per base (LiDAR-only: pp4, sd1, vn6; cross-modality: EfficientNet, ResNet).}

\label{Tab:vn4real}
 \resizebox{0.80\linewidth}{!}{%
\begin{tabular}{c|c|c|ccc|c}
\toprule
Stage-1 Scenarios & Interpreter &AP@0.5/0.7(\%)& $S_0$&$S_1$ & $S_2$&$S_{avg}$\\
\midrule
pp4-vn4 & pp4 & 84.1/65.2&0.8689&0.8235&0.7618&0.7947\\
\textbf{pp4-vn4} &\textbf{vn4}& \textbf{95.6/83.0}&\textbf{0.8638}&\textbf{0.8531}&\textbf{0.8208}&\textbf{0.8362}\\
pp4-vn4 &pp4 (s-2)&88.6/70.3&0.7737&0.7668&0.8121&0.7937\\
pp4-vn4 &sd1 (s-2)&85.0/65.8&0.8164&0.8346&0.7416&0.7788\\
pp4-vn4 &vn6 (s-2) &86.3/68.0&0.7290&0.7465&0.8232&0.7878\\
\midrule
\textbf{vn4-sd2} & \textbf{vn4 }& \textbf{95.6/84.0}&\textbf{0.9027}&\textbf{0.8273}&	\textbf{0.8816}&\textbf{0.8691}\\
vn4-sd2 &sd2 &80.9/60.4&0.8421&0.7324&0.8572&0.8194
\\
vn4-sd2& pp4 (s-2) &83.1/61.5&0.8440&0.7950&0.8109&0.8111\\
vn4-sd2 &sd1 (s-2)&78.9/60.6&0.8130&0.8354&0.8237&0.8255\\
vn4-sd2 &vn6 (s-2)&85.3/65.1&0.8000&0.7261&0.7944&0.7757\\
\midrule
\textbf{vn4-ResNet} &\textbf{ vn4} &\textbf{95.0/82.1}&\textbf{0.8626}&\textbf{0.8194}&\textbf{0.8772}&\textbf{0.8586}
\\
vn4-ResNet &ResNet &89.1/72.4&0.7732&0.7272&0.6711&0.7017\\
vn4-ResNet &EfficientNet (s-2)&88.6/71.9&0.9006&0.7671&0.6786&0.7356 \\
\midrule
pp4-EfficientNet & pp4&87.8/68.5&0.7588&0.8469&0.8255&0.8221
\\
pp4-EfficientNet & EfficientNet &86.3/68.1&0.7578&0.8040&0.8489&0.8231
\\
pp4-EfficientNet &ResNet (s-2)& 88.4/70.5&0.8191&0.8150&0.7920&0.8025\\
\midrule
\midrule
Average over &  all& 87.3/69.7&0.8204&0.7950&0.8013&0.8022\\
Minimum AP &all&78.9/60.6&0.8130&0.8354&0.8237&0.8255\\
\textbf{Average over}  &\textbf{pretrained vn4} &\textbf{95.4/83.0}&\textbf{0.8764}&\textbf{0.8333}&\textbf{0.8599}&\textbf{0.8546}\\
\bottomrule
\end{tabular}
}
\end{table*}
The detailed results for sd1 are shown in Table~\ref{Tab:sd1real}. Interpreters that were trained with sd1 (shown in bold) again achieve the highest average similarities ($S_{\mathrm{avg}}$) and the best detection performance. Interpreters pretrained on sd1 gain up to \textbf{+18.2\%} AP@0.5 and \textbf{+29.6\%} AP@0.7, and \textbf{+10.0\%} AP@0.5 and \textbf{+18.9\%} AP@0.7 over the average performance across all interpreters. The mean similarity between the sd1 domain-invariant features and the ego features for interpreters pretrained on sd1 exceeds the average over all interpreters by $0.09$ (with similarity threshold $\tau{=}0.87$).

\begin{table*}[t]
\centering
\caption{Detection performance (AP@0.5 / AP@0.7(\%)), feature similarity for scales $0$--$2$ ($S_0$--$S_2$) and their average ($S_{\mathrm{avg}}$) for real-time anonymous/unseen model selection with sd1. Heterogeneous bases: pp4-vn4, vn4-sd2, vn4-ResNet, pp4-EfficientNet with pp8 as ego. Stage-2 (s-2): new-agent adaptation per base (LiDAR-only: pp4, sd1, vn6; cross-modality: EfficientNet, ResNet).}

\label{Tab:sd1real}
 \resizebox{0.80\linewidth}{!}{%
\begin{tabular}{c|c|c|ccc|c}
\toprule
Stage-1 Scenarios & Interpreter &AP@0.5/0.7(\%)& $S_0$&$S_1$ & $S_2$&$S_{avg}$\\
\midrule
pp4-vn4 &pp4 & 87.0/65.3&0.7937&0.8237&0.8325&0.8245\\
pp4-vn4 &vn4& 83.0/59.3&0.7837&0.8306&0.8660&0.8441\\
pp4-vn4 &pp4 (s-2) &85.3/61.0&0.7544&0.7711&0.8465&0.8118\\
\textbf{pp4-vn4} &\textbf{sd1 (s-2)}&\textbf{95.7/85.0}&\textbf{0.9138}&\textbf{0.8538}&\textbf{0.9375}&\textbf{0.9102}\\
pp4-vn4 &vn6 (s-2) &79.8/58.1&0.7618&0.7880&0.9003&0.8484 \\
\midrule
vn4-sd2 & vn4 & 84.6/63.3&0.8035&0.8247&0.8775&0.8519
\\
vn4-sd2 &sd2 &80.4/60.0&0.8056&0.7567&0.8684&0.8275\\
vn4-sd2 &pp4 (s-2)&77.6/56.0&0.8273&0.7778&0.7837&0.7882\\
\textbf{vn4-sd2} &\textbf{sd1 (s-2)}&\textbf{95.9/86.3}&\textbf{0.8973}&\textbf{0.9018}&\textbf{0.8928}&\textbf{0.8960}\\
vn4-sd2 &vn6 (s-2)&81.6/63.8&0.8373&0.8169&0.8549&0.8416\\
\midrule
vn4-ResNet & vn4 &84.5/64.3&0.8511&0.7744&0.7641&0.7795\\
vn4-ResNet &ResNet &88.2/69.7&0.7659&0.8710&0.8832&0.8630\\
vn4-ResNet &EfficientNet (s-2)&88.3/68.7&0.7754&0.7165&0.5628&0.6371 \\
\midrule
pp4-EfficientNet & pp4&87.0/68.0&0.7785&0.8639&0.8015&0.8160\\
pp4-EfficientNet & EfficientNet &86.6/70.0&0.8508&0.7642&0.5953&0.6801\\
pp4-EfficientNet &ResNet (s-2)& 86.6/69.3&0.7846&0.7746&0.7699&0.7733\\
\midrule
\midrule
Average over &  all& 85.8/66.7&0.8115&0.8069&0.8148&0.8121\\
Minimum AP &all&77.6/56.0&0.8273&0.7778&0.7837&0.7882\\
\textbf{Average over}  &\textbf{pretrained sd1} &\textbf{95.8/85.6}&\textbf{0.9055}&\textbf{0.8778}&\textbf{0.9151}&\textbf{0.9031}\\
\bottomrule
\end{tabular}
}
\end{table*}


For pp6, the detailed results are shown in Table~\ref{Tab:pp6real}. Since pp6 is evaluated as an unseen or novel joining agent, the selected interpreter is compared with the average performance across all candidate interpreters. The top selected interpreter improves performance by \textbf{+4.5\%} AP@0.5 and \textbf{+5.5\%} AP@0.7 over the average performance across all interpreters. The mean similarity for the top selected interpreter exceeds the all-interpreter average by $0.08$, indicating that the learned domain-invariant similarity remains informative even for unseen agents.

\begin{table*}[t]
\centering
\caption{Detection performance (AP@0.5 / AP@0.7(\%)), feature similarity for scales $0$--$2$ ($S_0$--$S_2$) and their average ($S_{\mathrm{avg}}$) for real-time anonymous/unseen model selection with pp6. Heterogeneous bases: pp4-vn4, vn4-sd2, vn4-ResNet, pp4-EfficientNet with pp8 as ego. Stage-2 (s-2): new-agent adaptation per base (LiDAR-only: pp4, sd1, vn6; cross-modality: EfficientNet, ResNet).}
\label{Tab:pp6real}
\resizebox{0.80\linewidth}{!}{%
\begin{tabular}{c|c|c|ccc|c}
\toprule
Stage-1 Scenarios & Interpreter & AP@0.5/0.7(\%) & $S_0$ & $S_1$ & $S_2$ & $S_{\mathrm{avg}}$\\
\midrule
pp4-vn4 & pp4 & 80.3/61.2 & 0.8066 & 0.8171 & 0.7343 & 0.7683\\
pp4-vn4 & vn4 & 82.0/63.0 & 0.8423 & 0.8190 & 0.6960 & 0.7521\\
pp4-vn4 & sd1 (s-2) & 82.1/62.8 & 0.8251 & 0.7848 & 0.9076 & 0.8608\\
pp4-vn4 & pp4 (s-2) & 81.9/63.4 & 0.8259 & 0.8069 & 0.8878 & 0.8559\\
pp4-vn4 & vn6 (s-2) & 84.5/66.0 & 0.7920 & 0.7859 & 0.8757 & 0.8381\\
\midrule
vn4-sd2 & vn4 & 79.0/59.8 & 0.8401 & 0.8212 & 0.8726 & 0.8533\\
vn4-sd2 & sd2 & 81.6/61.7 & 0.7954 & 0.7822 & 0.8263 & 0.8093\\
vn4-sd2 & sd1 (s-2) & 79.1/54.9 & 0.8384 & 0.8468 & 0.8854 & 0.8676\\
vn4-sd2 & pp4 (s-2) & 79.4/59.3 & 0.8890 & 0.7768 & 0.7744 & 0.7915\\
vn4-sd2 & vn6 (s-2) & 82.0/63.1 & 0.8100 & 0.7824 & 0.8624 & 0.8320\\
\midrule
vn4-ResNet & vn4 & 87.1/69.7 & 0.7446 & 0.8172 & 0.8755 & 0.8401\\
vn4-ResNet & ResNet & 84.6/66.3 & 0.7897 & 0.7821 & 0.8038 & 0.7956\\
vn4-ResNet & EfficientNet (s-2) & 87.9/70.5 & 0.7894 & 0.7893 & 0.4684 & 0.6059\\
\midrule
pp4-EfficientNet & pp4 & 87.8/70.7 & 0.8349 & 0.7255 & 0.5390 & 0.6346\\
pp4-EfficientNet & EfficientNet & 85.5/68.4 & 0.7692 & 0.8576 & 0.8568 & 0.8445\\
\textbf{pp4-EfficientNet} & \textbf{ResNet (s-2)} & \textbf{87.8/69.9} & \textbf{0.7821} & \textbf{0.8535} & \textbf{0.9259} & \textbf{0.8847}\\
\midrule
\midrule
Average over & all & 83.3/64.4 & 0.8109 & 0.8030 & 0.7994 & 0.8021\\
\textbf{Top} & \textbf{selected} & \textbf{87.8/69.9} & \textbf{0.7821} & \textbf{0.8535} & \textbf{0.9259} & \textbf{0.8847}\\
\bottomrule
\end{tabular}
}
\end{table*}

\subsection{V2XSet}
\label{sec:real-V2XSet}
We also evaluated our real-time interpreter selection process on V2XSet. Following Section~\ref{sec:real-opv2v}, we test the anonymous selection protocol using pp4, vn4, and pp6 as anonymous neighbors. 
The trends match those observed on OPV2V: the cosine similarity-based selection reliably identifies the most suitable interpreter among the pretrained models, improving AP over random interpreter selection while maintaining a small selection overhead on the ego side.

For pp4, the detailed results are shown in Table~\ref{Tab:pp4real1}. Interpreters that were trained with pp4 (shown in bold) achieve the highest average similarities ($S_{\mathrm{avg}}$) and the best detection performance. Interpreters pretrained on pp4 gain \textbf{+8.0\%} AP@0.5 and \textbf{+12.8\%} AP@0.7 over the average performance across all interpreters (i.e., under random selection). The mean similarity between the pp4 domain-invariant features and the ego features for interpreters pretrained on pp4 exceeds the all-interpreter average by $0.05$.

\begin{table*}[t]
\centering
\caption{Detection performance (AP@0.5 / AP@0.7, \%), feature similarity for scales $0$--$2$ ($S_0$--$S_2$) and their average ($S_{\mathrm{avg}}$) for real-time anonymous model selection with pp4 on V2XSet. Heterogeneous bases: pp4-vn4 and vn4-sd2, with pp8 as ego. Stage-2 (s-2): new-agent adaptation per base.}
\label{Tab:pp4real1}
\resizebox{0.80\linewidth}{!}{%
\begin{tabular}{c|c|c|ccc|c}
\toprule
Stage-1 Scenarios & Interpreter & AP@0.5/0.7(\%) & $S_0$ & $S_1$ & $S_2$ & $S_{\mathrm{avg}}$\\
\midrule
\textbf{pp4-vn4} & \textbf{pp4} & \textbf{91.3/71.2} & \textbf{0.8654} & \textbf{0.8222} & \textbf{0.9430} & \textbf{0.8974} \\
pp4-vn4 & vn4 & 80.9/54.7 & 0.8040 & 0.7812 & 0.9278 & 0.8682 \\
pp4-vn4 & sd1 (s-2) & 79.7/52.3 & 0.7828 & 0.7531 & 0.8161 & 0.7933 \\
\textbf{pp4-vn4} & \textbf{pp4 (s-2)} & \textbf{91.2/71.7} & \textbf{0.8307} & \textbf{0.8241} & \textbf{0.8057} & \textbf{0.8145} \\
pp4-vn4 & vn6 (s-2) & 77.2/51.2 & 0.8248 & 0.7791 & 0.8933 & 0.8509 \\
\midrule
vn4-sd2 & vn4 & 77.0/51.3 & 0.8513 & 0.7982 & 0.8178 & 0.8170 \\
vn4-sd2 & sd2 & 79.9/52.7 & 0.7596 & 0.7645 & 0.6513 & 0.6991 \\
vn4-sd2 & sd1 (s-2) & 85.0/57.6 & 0.7645 & 0.7860 & 0.7517 & 0.7633 \\
\textbf{vn4-sd2} & \textbf{pp4 (s-2)} & \textbf{91.4/72.3} & \textbf{0.8790} & \textbf{0.8449} & \textbf{0.8908} & \textbf{0.8760} \\
vn4-sd2 & vn6 (s-2) & 79.1/53.6 & 0.7650 & 0.7246 & 0.7719 & 0.7574 \\
\midrule
\midrule
Average over & all & 83.3/58.9 & 0.8127 & 0.7878 & 0.8269 & 0.8137 \\
\textbf{Average over} & \textbf{pretrained pp4} & \textbf{91.3/71.7} & \textbf{0.8584} & \textbf{0.8304} & \textbf{0.8798} & \textbf{0.8626} \\
\bottomrule
\end{tabular}
}
\end{table*}

For vn4, the detailed results are shown in Table~\ref{Tab:vn4real1}. Interpreters that were trained with vn4 (shown in bold) achieve the highest average similarities ($S_{\mathrm{avg}}$) and the best detection performance. Interpreters pretrained on vn4 gain \textbf{+7.1\%} AP@0.5 and \textbf{+12.9\%} AP@0.7 over the average performance across all interpreters (i.e., under random selection). The mean similarity between the vn4 domain-invariant features and the ego features for interpreters pretrained on vn4 exceeds the average over all interpreters by $0.06$.

\begin{table*}[!t]
\centering
\caption{Detection performance (AP@0.5 / AP@0.7, \%), feature similarity for scales $0$--$2$ ($S_0$--$S_2$) and their average ($S_{\mathrm{avg}}$) for real-time anonymous model selection with vn4 on V2XSet. Heterogeneous bases: pp4-vn4 and vn4-sd2, with pp8 as ego. Stage-2 (s-2): new-agent adaptation per base.}
\label{Tab:vn4real1}
\resizebox{0.80\linewidth}{!}{%
\begin{tabular}{c|c|c|ccc|c}
\toprule
Stage-1 Scenarios & Interpreter & AP@0.5/0.7(\%) & $S_0$ & $S_1$ & $S_2$ & $S_{\mathrm{avg}}$\\
\midrule
\textbf{pp4-vn4} & \textbf{pp4} & \textbf{85.4/59.4} & \textbf{0.7532} & \textbf{0.7679} & \textbf{0.8854} & \textbf{0.8329} \\
\textbf{pp4-vn4} & \textbf{vn4} & \textbf{91.3/72.5} & \textbf{0.8008} & \textbf{0.8058} & \textbf{0.9369} & \textbf{0.8800} \\
pp4-vn4 & sd1 (s-2) & 83.6/57.5 & 0.8323 & 0.7769 & 0.7696 & 0.7806 \\
pp4-vn4 & pp4 (s-2) & 82.5/56.9 & 0.8071 & 0.7241 & 0.7515 & 0.7516 \\
pp4-vn4 & vn6 (s-2) & 79.1/50.1 & 0.7701 & 0.7322 & 0.8367 & 0.7973 \\
\midrule
\textbf{vn4-sd2} & \textbf{vn4} & \textbf{91.2/71.9} & \textbf{0.8607} & \textbf{0.8278} & \textbf{0.8409} & \textbf{0.8400} \\
vn4-sd2 & sd2 & 82.6/55.2 & 0.7741 & 0.7933 & 0.8113 & 0.8008 \\
vn4-sd2 & sd1 (s-2) & 83.0/57.5 & 0.7723 & 0.8105 & 0.8168 & 0.8086 \\
vn4-sd2 & pp4 (s-2) & 85.1/58.0 & 0.8062 & 0.7376 & 0.6190 & 0.6796 \\
vn4-sd2 & vn6 (s-2) & 78.7/53.9 & 0.8456 & 0.8183 & 0.8297 & 0.8287 \\
\midrule
\midrule
Average over & all & 84.2/59.3 & 0.8022 & 0.7794 & 0.8098 & 0.8000 \\
\textbf{Average over} & \textbf{pretrained vn4} & \textbf{91.3/72.2} & \textbf{0.8308} & \textbf{0.8168} & \textbf{0.8889} & \textbf{0.8600} \\
\bottomrule
\end{tabular}
}
\end{table*}

For pp6, the detailed results are shown in Table~\ref{Tab:pp6real1}. Since pp6 is evaluated as an unseen or less-matched anonymous joining agent, the selected interpreter is compared with the average performance across all candidate interpreters. The top selected interpreter improves performance by \textbf{+2.5\%} AP@0.5 and \textbf{+2.1\%} AP@0.7 over the average performance across all interpreters. The mean similarity for the top selected interpreter exceeds the all-interpreter average by $0.10$, showing that the similarity score remains informative for V2XSet dataset.

\begin{table*}[t]
\centering
\caption{Detection performance (AP@0.5 / AP@0.7, \%), feature similarity for scales $0$--$2$ ($S_0$--$S_2$) and their average ($S_{\mathrm{avg}}$) for real-time anonymous model selection with pp6 on V2XSet. Heterogeneous bases: pp4-vn4 and vn4-sd2, with pp8 as ego. Stage-2 (s-2): new-agent adaptation per base.}
\label{Tab:pp6real1}
\resizebox{0.80\linewidth}{!}{%
\begin{tabular}{c|c|c|ccc|c}
\toprule
Stage-1 Scenarios & Interpreter & AP@0.5/0.7(\%) & $S_0$ & $S_1$ & $S_2$ & $S_{\mathrm{avg}}$\\
\midrule
pp4-vn4 & pp4 & 77.4/50.8 & 0.8541 & 0.7983 & 0.8471 & 0.8342 \\
pp4-vn4 & vn4 & 77.2/50.0 & 0.7986 & 0.7838 & 0.8710 & 0.8357 \\

pp4-vn4 & sd1 (s-2) & 73.3/46.0 & 0.7688 & 0.6741 & 0.3284 & 0.4901 \\
pp4-vn4 & pp4 (s-2) & 76.1/50.0 & 0.8386 & 0.7839 & 0.8479 & 0.8283 \\
\textbf{pp4-vn4} & \textbf{vn6 (s-2)} & \textbf{79.4/53.1} & \textbf{0.8144} & \textbf{0.8590} & \textbf{0.8809} & \textbf{0.8651} \\
\midrule
vn4-sd2 & vn4 & 74.2/50.0 & 0.8378 & 0.7944 & 0.8058 & 0.8071 \\
vn4-sd2 & sd2 & 76.0/51.0 & 0.7623 & 0.7645 & 0.6943 & 0.7241 \\
vn4-sd2 & sd1 (s-2) & 77.3/51.9 & 0.7975 & 0.7815 & 0.7791 & 0.7824 \\
vn4-sd2 & pp4 (s-2) & 78.6/53.6 & 0.7639 & 0.7604 & 0.7334 & 0.7455 \\
vn4-sd2 & vn6 (s-2) & 79.1/53.7 & 0.7740 & 0.7422 & 0.7301 & 0.7398 \\
\midrule
\midrule
Average over & all & 76.9/51.0 & 0.8010 & 0.7742 & 0.7518 & 0.7652 \\
\textbf{Top} & \textbf{selected} & \textbf{79.4/53.1} & \textbf{0.8144} & \textbf{0.8590} & \textbf{0.8809} & \textbf{0.8651} \\
\bottomrule
\end{tabular}
}
\end{table*}

\subsection{DAIR-V2X}
\label{sec:real-dairv2x}

We also evaluated our real-time interpreter selection process on DAIR-V2X. Following Sections~\ref{sec:real-opv2v} and~\ref{sec:real-V2XSet}, we test the anonymous selection protocol using pp4, vn4, and pp6 as anonymous neighbors. Similar trends are observed under this real-world benchmark: cosine similarity-based selection reliably identifies the most suitable interpreter among pretrained models, improving AP over random interpreter selection while maintaining low selection overhead on the ego side.

For pp4, the detailed results are shown in Table~\ref{Tab:pp4real1_new}. Interpreters pretrained with pp4 (shown in bold) achieve the highest average similarities ($S_{\mathrm{avg}}$) and the best detection performance. Interpreters pretrained on pp4 gain \textbf{+2.8\%} AP@0.5 and \textbf{+1.5\%} AP@0.7 over the average performance across all interpreters (i.e., under random selection). The mean similarity between the pp4 domain-invariant features and the ego features for interpreters pretrained on pp4 exceeds the all-interpreter average by $0.07$ (with similarity threshold $\tau{=}0.71$).

\begin{table*}[t]
\centering 
\caption{Detection performance (AP@0.5 / AP@0.7, \%), feature similarity for scales $0$--$2$ ($S_0$--$S_2$) and their average ($S_{\mathrm{avg}}$) for real-time anonymous model selection with pp4 on DAIR-V2X. Heterogeneous bases: pp4-vn4 and vn4-sd2, with pp8 as ego. Stage-2 (s-2): new-agent adaptation per base.}
\label{Tab:pp4real1_new}
\resizebox{0.80\linewidth}{!}{%
\begin{tabular}{c|c|c|ccc|c}
\toprule
Stage-1 Scenarios & Interpreter & AP@0.5/0.7(\%) & $S_0$ & $S_1$ & $S_2$ & $S_{\mathrm{avg}}$\\
\midrule
\textbf{pp4-vn4} & \textbf{pp4} & \textbf{71.7/45.2} & \textbf{0.8829} & \textbf{0.8330} & \textbf{0.6200} & \textbf{0.7184} \\
pp4-vn4 & vn4 & 69.0/43.1 & 0.8036 & 0.7393 & 0.4306 & 0.5721 \\
pp4-vn4 & sd1 (s-2) & 67.7/43.0 & 0.7806 & 0.8345 & 0.4343 & 0.5981 \\
\textbf{pp4-vn4} & \textbf{pp4 (s-2)} & \textbf{71.6/45.0} & \textbf{0.8809} & \textbf{0.8449} & \textbf{0.6478} & \textbf{0.7374} \\
pp4-vn4 & vn6 (s-2) & 67.2/42.5 & 0.6534 & 0.7607 & 0.5884 & 0.6469 \\
\midrule
vn4-sd2 & vn4 & 67.1/42.4 & 0.7761 & 0.7559 & 0.5224 & 0.6254 \\
vn4-sd2 & sd2 & 67.3/42.6 & 0.8110 & 0.7882 & 0.5984 & 0.6830 \\
vn4-sd2 & sd1 (s-2) & 68.0/43.1 & 0.8745 & 0.8226 & 0.5597 & 0.6798 \\
\textbf{vn4-sd2} & \textbf{pp4 (s-2)} & \textbf{71.6/44.9} & \textbf{0.8441} & \textbf{0.8488} & \textbf{0.6863} & \textbf{0.7553} \\
vn4-sd2 & vn6 (s-2) & 66.9/42.9 & 0.8312 & 0.7303 & 0.5959 & 0.6679 \\
\midrule
\midrule
Average over & all & 68.8/43.5 & 0.8138 & 0.7958 & 0.5684 & 0.6684 \\
\textbf{Average over} & \textbf{pretrained pp4} & \textbf{71.6/45.0} & \textbf{0.8693} & \textbf{0.8422} & \textbf{0.6514} & \textbf{0.7370} \\
\bottomrule
\end{tabular}
}
\end{table*}

For vn4, the detailed results are shown in Table~\ref{Tab:vn4real1_new}. Interpreters pretrained with vn4 (shown in bold) achieve the highest average similarities ($S_{\mathrm{avg}}$) and the best detection performance. Interpreters pretrained on vn4 gain \textbf{+2.9\%} AP@0.5 and \textbf{+1.7\%} AP@0.7 over the average performance across all interpreters (i.e., under random selection). The mean similarity between the vn4 domain-invariant features and the ego features for interpreters pretrained on vn4 exceeds the all-interpreter average by $0.10$ (with similarity threshold $\tau{=}0.74$).

\begin{table*}[t]
\centering
\caption{Detection performance (AP@0.5 / AP@0.7, \%), feature similarity for scales $0$--$2$ ($S_0$--$S_2$) and their average ($S_{\mathrm{avg}}$) for real-time anonymous model selection with vn4 on DAIR-V2X. Heterogeneous bases: pp4-vn4 and vn4-sd2, with pp8 as ego. Stage-2 (s-2): new-agent adaptation per base.}
\label{Tab:vn4real1_new}
\resizebox{0.80\linewidth}{!}{%
\begin{tabular}{c|c|c|ccc|c}
\toprule
Stage-1 Scenarios & Interpreter & AP@0.5/0.7(\%) & $S_0$ & $S_1$ & $S_2$ & $S_{\mathrm{avg}}$\\
\midrule
pp4-vn4 & pp4 & 70.2/43.8 & 0.7915 & 0.7674 & 0.3605 & 0.5383 \\
\textbf{pp4-vn4} & \textbf{vn4} & \textbf{72.5/45.7} & \textbf{0.8485} & \textbf{0.8381} & \textbf{0.6654} & \textbf{0.7409} \\
pp4-vn4 & sd1 (s-2) & 68.9/43.5 & 0.8057 & 0.7935 & 0.4904 & 0.6220 \\
pp4-vn4 & pp4 (s-2) & 70.2/43.7 & 0.7964 & 0.7187 & 0.4834 & 0.5953 \\
pp4-vn4 & vn6 (s-2) & 67.6/42.3 & 0.6160 & 0.7935 & 0.5648 & 0.6375 \\
\midrule
\textbf{vn4-sd2} & \textbf{vn4} & \textbf{72.2/45.1} & \textbf{0.8270} & \textbf{0.8530} & \textbf{0.6970} & \textbf{0.7601} \\
vn4-sd2 & sd2 & 67.8/43.0 & 0.8678 & 0.7497 & 0.5495 & 0.6522 \\
vn4-sd2 & sd1 (s-2) & 68.1/43.2 & 0.7933 & 0.8191 & 0.4232 & 0.5892 \\
vn4-sd2 & pp4 (s-2) & 70.5/44.0 & 0.8023 & 0.8067 & 0.6683 & 0.7270 \\
vn4-sd2 & vn6 (s-2) & 66.9/42.6 & 0.8476 & 0.7521 & 0.5469 & 0.6485 \\
\midrule
\midrule
Average over & all & 69.5/43.7 & 0.7996 & 0.7892 & 0.5449 & 0.6511 \\
\textbf{Average over} & \textbf{pretrained vn4} & \textbf{72.4/45.4} & \textbf{0.8378} & \textbf{0.8456} & \textbf{0.6812} & \textbf{0.7505} \\
\bottomrule
\end{tabular}
}
\end{table*}

For pp6, the detailed results are shown in Table~\ref{Tab:pp6real1_new}. Since pp6 is evaluated as an unseen or less-matched anonymous joining agent, the selected interpreter is compared with the average performance across all candidate interpreters. The top selected interpreter improves performance by \textbf{+0.3\%} AP@0.5 and \textbf{+0.6\%} AP@0.7 over the average performance across all interpreters. The mean similarity for the top selected interpreter exceeds the all-interpreter average by $0.06$. Although the detection gain is modest, the similarity score follows the same trend observed on OPV2V and V2XSet, suggesting that \textsc{PEARL}'s feature-based assignment remains consistent under real-world vehicle-infrastructure domain shifts.

\begin{table*}[t]
\centering
\caption{Detection performance (AP@0.5 / AP@0.7, \%), feature similarity for scales $0$--$2$ ($S_0$--$S_2$) and their average ($S_{\mathrm{avg}}$) for real-time anonymous model selection with pp6 on DAIR-V2X. Heterogeneous bases: pp4-vn4 and vn4-sd2, with pp8 as ego. Stage-2 (s-2): new-agent adaptation per base.}
\label{Tab:pp6real1_new}
\resizebox{0.80\linewidth}{!}{%
\begin{tabular}{c|c|c|ccc|c}
\toprule
Stage-1 Scenarios & Interpreter & AP@0.5/0.7(\%) & $S_0$ & $S_1$ & $S_2$ & $S_{\mathrm{avg}}$\\
\midrule
\textbf{pp4-vn4} & \textbf{pp4} & \textbf{67.5/43.1} & \textbf{0.8664} & \textbf{0.7832} & \textbf{0.7099} & \textbf{0.7532}\\
pp4-vn4 & vn4 & 67.0/42.1 & 0.7712 & 0.7463 & 0.6837 & 0.7142 \\
pp4-vn4 & sd1 (s-2) & 67.5/42.7 & 0.7899 & 0.8146 & 0.7043 & 0.7480 \\
pp4-vn4 & pp4 (s-2) & 67.4/42.7 & 0.8399 & 0.8166 & 0.6149 & 0.7046\\
pp4-vn4 & vn6 (s-2) & 67.0/42.1 & 0.7195 & 0.8248 & 0.6038 & 0.6835\\
\midrule
vn4-sd2 & vn4 & 67.1/42.3 & 0.7776 & 0.7578 & 0.5173 & 0.6232\\
vn4-sd2 & sd2 & 67.0/42.0 & 0.8198 & 0.7797 & 0.6327 & 0.7014\\
vn4-sd2 & sd1 (s-2) & 66.9/42.4 & 0.8499 & 0.7941 & 0.5090 & 0.6392\\
vn4-sd2 & pp4 (s-2) & 67.4/42.5 & 0.7472 & 0.7667 & 0.7304 & 0.7432\\
vn4-sd2 & vn6 (s-2) & 67.4/42.9 & 0.8422 & 0.7493 & 0.5460 & 0.6464\\
\midrule
\midrule
Average over & all & 67.2/42.5 & 0.8023 & 0.7833 & 0.6252 & 0.6957\\
\textbf{Top} & \textbf{selected} & \textbf{67.5/43.1} & \textbf{0.8664} & \textbf{0.7832} & \textbf{0.7099} & \textbf{0.7532}\\
\bottomrule
\end{tabular}
}
\end{table*}
\subsection{Real-Time Complexity}
\label{sec:time}
Detailed timing breakdowns are reported in Table~\ref{Tab:time}. To evaluate \textsc{PEARL}'s real-time complexity, we measure the per-sample GPU time for interpreter selection and collaborative detection, where each sample contains both an ego and a neighbor agent. On average, the Pipeline-2 real-time selection step takes \textbf{1.67 ms}, and the MS extractor takes \textbf{47.0 ms} per sample. This yields an end-to-end selection time of \textbf{48.67 ms} per sample.
Here, the Pipeline-2 latency (\textbf{1.67 ms}) corresponds to the similarity-based interpreter selection in the domain-invariant feature space, while the additional cost arises from applying the multi-scale (MS) feature extractor for each candidate interpreter. The fastest interpreters and heterogeneous bases are those based on \texttt{pp4} (PointPillar), while the slowest ones involve camera-based backbones. Importantly, this selection overhead is incurred only on the ego side and does not require querying or executing any additional encoders, which makes the design practical for deployment.

After the ego connects with the neighbor, it first performs the end-to-end selection process to choose a suitable pretrained detection interpreter, and subsequently uses only the selected interpreter for cooperative detection.
\begin{table*}[t]
\centering
\caption{Total Pipeline-2 and MS extractor computation time for Stage-2 interpreters (in milliseconds). Heterogeneous bases: pp4-vn4, vn4-sd2, vn4-ResNet, pp4-EfficientNet with pp8 as ego. Stage-2: new-agent adaptation per base (LiDAR-only: pp4, sd1, vn6; cross-modality: EfficientNet, ResNet).}

\label{Tab:time}
 \resizebox{0.80\linewidth}{!}{%
\begin{tabular}{c|c|cc|c}
\toprule
Stage-1 Scenarios & Interpreter & Pipeline-2&MS extractor&Pipeline-2+MS extractor\\
\midrule
pp4-vn4 &pp4 &1.60&27.7&29.30\\
pp4-vn4 &sd1  &1.64&58.0&59.64\\
pp4-vn4 &vn6  &1.65&50.5&52.15\\
\midrule
vn4-sd2 &pp4 &1.61&19.9&21.51\\
vn4-sd2 &sd1 &1.65&59.3&60.95\\
vn4-sd2 &vn6 &1.67&54.5&56.17\\
\midrule
vn4-ResNet &EfficientNet &1.76&53.6&55.36
\\
\midrule
pp4-EfficientNet &ResNet&1.80&52.5&54.30\\
\midrule
\midrule
Average over &all&1.67&47.0&48.67\\
\bottomrule
\end{tabular}
}
\end{table*}

\subsection{Detailed Cross-Dataset Interpreter Selection}
\label{sec:cross-dataset-detailed}

We further report the full cross-dataset interpreter-selection results for the \texttt{pp4} encoder trained on OPV2V when the candidate interpreter pool is trained on V2XSet. Table~\ref{Tab:pp4_cross_full} lists detection performance and similarity scores for all available interpreters. The top-3 selected interpreters, determined solely by the highest average cosine similarity $S_{\mathrm{avg}}$, are used in Table~\ref{Tab:pp4_cross}.

Importantly, even under cross-dataset domain shift, the interpreter with the highest similarity score is also the best available interpreter in terms of detection performance. Specifically, the \texttt{sd1} Stage-2 interpreter trained from the \texttt{vn4,sd2} heterogeneous base achieves the highest $S_{\mathrm{avg}}=0.8438$ and also the best detection accuracy, reaching 89.8\% AP@0.5 and 63.4\% AP@0.7. This result confirms that \textsc{PEARL}'s similarity score remains predictive of interpreter quality even when the joining agent and interpreter pool come from different datasets.

Notably, the best cross-dataset interpreter is not the interpreter associated with the same nominal metadata as the joining agent. Although the joining agent uses a \texttt{pp4} encoder, the best available V2XSet-trained interpreter is the \texttt{sd1} Stage-2 interpreter rather than a \texttt{pp4} interpreter. This observation further motivates anonymous, feature-based interpreter selection: even if a neighboring agent discloses its model or sensor metadata, that metadata may not identify the most compatible interpreter under dataset-level distribution shift.

\begin{table*}[t]
\centering
\caption{Cross-dataset detection performance (AP@0.5 / AP@0.7, \%), feature similarity for scales $0$--$2$ ($S_0$--$S_2$), and their average ($S_{\mathrm{avg}}$) for the OPV2V-trained pp4 encoder when selecting among interpreters trained on V2XSet. The top-3 interpreters selected by the highest $S_{\mathrm{avg}}$ are highlighted in bold.}
\label{Tab:pp4_cross_full}
\resizebox{0.80\linewidth}{!}{%
\begin{tabular}{c|c|c|ccc|c}
\toprule
Stage-1 Scenarios & Interpreter & AP@0.5/0.7(\%) & $S_0$ & $S_1$ & $S_2$ & $S_{\mathrm{avg}}$\\
\midrule
pp4-vn4 & vn4 & 82.6/58.1 & 0.8131 & 0.7760 & 0.8486 & 0.8228 \\
\textbf{pp4-vn4} & \textbf{pp4} & \textbf{84.9/57.4} & \textbf{0.8513} & \textbf{0.7931} & \textbf{0.8592} & \textbf{0.8392} \\
pp4-vn4 & sd1 (s-2) & 82.8/55.5 & 0.8417 & 0.7636 & 0.6842 & 0.7294 \\
\textbf{pp4-vn4} & \textbf{pp4 (s-2)} & \textbf{88.5/60.5} & \textbf{0.8047} & \textbf{0.7878} & \textbf{0.8802} & \textbf{0.8430} \\
pp4-vn4 & vn6 (s-2) & 85.1/58.0 & 0.7887 & 0.8169 & 0.7035 & 0.7481 \\
\midrule
vn4-sd2 & vn4 & 79.3/54.2 & 0.8624 & 0.8244 & 0.7365 & 0.7796 \\
vn4-sd2 & sd2 & 84.6/55.6 & 0.7431 & 0.7631 & 0.6488 & 0.6949 \\
\textbf{vn4-sd2} & \textbf{sd1 (s-2)} & \textbf{89.8/63.4} & \textbf{0.8083} & \textbf{0.7811} & \textbf{0.8840} & \textbf{0.8438} \\
vn4-sd2 & pp4 (s-2) & 85.2/59.5 & 0.8613 & 0.7652 & 0.5753 & 0.6704 \\
vn4-sd2 & vn6 (s-2) & 76.1/51.4 & 0.8437 & 0.7971 & 0.8086 & 0.8103 \\
\midrule
\midrule
Average over & all & 83.9/57.4 & 0.8218 & 0.7868 & 0.7629 & 0.7782 \\
\textbf{Average over} & \textbf{Top-3 selected} & \textbf{87.7/60.4} & \textbf{0.8214} & \textbf{0.7873} & \textbf{0.8745} & \textbf{0.8420} \\
\bottomrule
\end{tabular}}
\end{table*}

\clearpage
\balance
\section{Additional Ablation Study}
\label{sec:addabl}
\subsection{Ego BEV Feature and Detection Head Size}
We further investigate the effect of the ego BEV feature size and the number of channels in the detection head on detection performance. As shown in Table~\ref{tab:ablmodel3}, we use two variations of the ego BEV feature size $(C\times H \times W)$: $64 \times 64 \times 128$ and $64 \times 128 \times 256$, and two variations for the detection head channels: 256 and 384 channels. The configuration with an ego feature size of $64 \times 64 \times 128$ and a 384-channel detection head is used in the main paper and the appendix results. The ego BEV features are used by the compressor, the multi-scale extractor, and the first scale of the Pipeline-1 and Pipeline-2 components, i.e., \textsc{LWSD}, the foreground estimator, channel cross-attention, spatial attention, \textsc{LWDDI}, and the domain classifier. 

The number of trainable parameters for the $64 \times 64 \times 128$ and $64 \times 128 \times 256$ ego feature sizes is $12.4$ M and $16.7$ M in Stage-1, respectively. Our default configuration (first row) outperforms the configuration with a 256-channel detection head and ego feature size $64 \times 64 \times 128$ by 0.8\% AP@0.5 on average for both Stage-1 and Stage-2, and by 0.4/0.8\% AP@0.7 for Stage-1/Stage-2 on average. It also outperforms the configuration with a 256-channel detection head and ego feature size $64 \times 128 \times 256$ by 0.5/0.4\% AP@0.5 for Stage-1/Stage-2 on average, and by 0.2/0.3\% AP@0.7 for Stage-1/Stage-2 on average. 

On the other hand, the configuration with a 384-channel detection head and ego feature size $64 \times 128 \times 256$ yields slightly lower AP@0.5 (by 0.2/0.4\% for Stage-1/Stage-2), while improving AP@0.7 by 5.0/2.9\% for Stage-1/Stage-2 when compared to our default setting, at the cost of an additional $4.3$ M trainable parameters. Since our focus is on a real-time solution that is both fast and reliable, we choose $64 \times 64 \times 128$ as the ego BEV feature size and 384 channels in the detection head as a balanced detection performance/speed trade-off. We use this configuration for all real-time implementation and complexity experiments.

\begin{table}[t]
\centering
\caption{Effect of ego BEV feature size and detection head number of channels on detection performance. We use two variations of the ego BEV feature size $(C\times H \times W)$: $64 \times 64 \times 128$ and $64 \times 128 \times 256$, and two variations for the detection head number of channels: 256 and 384. The configuration with ego feature size \underline{ $64 \times 64 \times 128$ } and a \underline{384}-channel detection head is used for all previous results. We report AP@0.5 / AP@0.7.}
\label{tab:ablmodel3}
\resizebox{\linewidth}{!}{%
    \begin{tabular}{c|cccc}
    \toprule
         BEV size, Detection head& pp4-vn4 & pp4& sd1 & vn6\\
        \midrule
        \underline{$64\times 64 \times 128$},\: \underline{384} &\textbf{95.5}/83.6&\textbf{95.5}/83.7&\textbf{95.7}/85.1&\textbf{90.7}/73.7\\
        $64\times128 \times 256$,\:384 & 95.3/\textbf{88.6} & 95.2/\textbf{86.4}&95.5/\textbf{87.7}&90.1/\textbf{77.1}\\
        $64 \times 64\times128 $, \:256&94.7/83.2&94.6/82.9&94.8/84.2&90.0/73.1\\
        $64\times128 \times 256$,\: 256 & 95.0/83.4&95.1/83.4&95.4/84.8&90.2/73.3\\        
        \bottomrule
    \end{tabular}
    }
\end{table}

\subsection{LWDDI Ablation Study}

Table~\ref{tab:Rub-abl} presents an ablation study of Pipeline-2, evaluating the cosine similarity used for interpreter selection under different design choices, including extractor architecture, PARAFAC prompt rank, and ego BEV feature resolution. The reported similarity is computed in the \textsc{LWDDI} domain-invariant feature space and corresponds to the compatibility score used during runtime selection.

Overall, the default \textsc{PEARL} configuration (multi-scale extractor + PARAFAC rank \texttt{[8,16,32]} + ego BEV size $64{\times}64{\times}128)$ achieves the highest and most consistent similarity across heterogeneous agent types, indicating stronger feature alignment and more reliable interpreter selection. Although the dense 3D prompt achieves a slightly higher similarity for vn6, the default configuration provides the highest average similarity and a substantially lower prompt-parameter count. Removing the multi-scale design or enforcing a shared extractor reduces similarity across all settings, suggesting weaker alignment of domain-invariant representations. Finally, increasing the ego BEV resolution introduces greater alignment difficulty, leading to lower similarity scores in this setup. This highlights the importance of maintaining a compact and well-structured feature space for stable domain-invariant alignment.
\begin{table}[t]
\centering
\caption{Ablation study of Pipeline-2 (\textsc{LWDDI}) on OPV2V, reporting cosine similarity for interpreter selection. Each variant modifies one component of the default \textsc{PEARL} configuration: extractor design, prompt rank, and feature resolution.} 
\label{tab:Rub-abl}
\footnotesize
\setlength{\tabcolsep}{3pt}
\resizebox{\columnwidth}{!}{%
\begin{tabular}{lcccc}
\toprule
\textbf{Variant} & \textbf{pp4-vn4} & \textbf{pp4} & \textbf{sd1} & \textbf{vn6} \\
\midrule
\textbf{PEARL (default)} & \textbf{0.8685} & \textbf{0.8797} & \textbf{0.8974} & 0.8718 \\
Single-scale LWDDI (--MS) & 0.8260 & 0.8376 & 0.8413 & 0.8363 \\
Shared MS extractor & 0.8362 & 0.8285 & 0.8393 & 0.8311 \\
Dense 3D prompt & 0.8558 & 0.8612 & 0.8908 &\textbf{ 0.8758} \\
PARAFAC rank \texttt{[8,8,8]} & 0.8549 & 0.8258 & 0.8466 & 0.8659 \\
PARAFAC rank \texttt{[16,16,16]} & 0.8471 & 0.8624 & 0.8685 & 0.8586 \\
PARAFAC rank \texttt{[32,32,32]} & 0.8245 & 0.8408 & 0.8765 & 0.8686 \\
Ego BEV size $64{\times}128{\times}256$ & 0.8221 & 0.8352 & 0.8674 & 0.8083 \\
\bottomrule
\end{tabular}%
}
\end{table}
\end{document}